%% file: sn-article.tex
\documentclass[pdflatex,sn-mathphys-num, iicol]{sn-jnl}

\usepackage{adjustbox} 
\usepackage{subcaption} 
\usepackage{tikz} 
\usepackage[edges]{forest} 
\definecolor{hidden-draw}{RGB}{20,68,106} 
\definecolor{hidden-pink}{RGB}{255,245,247}

\usepackage{graphicx}%
\usepackage{multirow}%
\usepackage{amsmath,amssymb,amsfonts}%
\usepackage{amsthm}%
\usepackage{mathrsfs}%
\usepackage[title]{appendix}%
\usepackage{textcomp}%
\usepackage{manyfoot}%
\usepackage{booktabs}%
\usepackage{algorithm}%
\usepackage{algorithmicx}%
\usepackage{algpseudocode}%
\usepackage{listings}%
\usepackage{natbib}
\usepackage{booktabs}
\usepackage{multirow}
\usepackage{colortbl}
\usepackage{xcolor}

\newcommand{\etal}{\textit{et al.}}

\theoremstyle{thmstyleone}%
\theoremstyle{thmstyletwo}%

\theoremstyle{thmstylethree}%

\begin{document}

\title[Towards Generalized Range-View LiDAR Segmentation\\ in Adverse Weather]{Towards Generalized Range-View LiDAR Segmentation\\ in Adverse Weather}


\author{~~~~~~~~~~Longyu Yang$^1$,~~~Lu Zhang$^2$,~~~Jun Liu$^3$,~~~Yap-Peng Tan$^4$,\\Hengtao Shen$^{1,5}$,~~~~Xiaofeng Zhu$^1$,~~~Ping Hu$^{1*}$}
\email{chinahuping@gmail.com}

\affil{\centering $^1$University of Electronic Science and Technology of China\\
        $^2$Dalian University of Technology\\
        $^3$Lancaster University\\
        $^4$Nanyang Technological University\\
        $^5$Tongji University}

\abstract{LiDAR segmentation has emerged as an important task to enrich scene perception and understanding. Range-view-based methods have gained popularity due to their high computational efficiency and compatibility with real-time deployment. However, their generalized performance under adverse weather conditions remains underexplored, limiting their reliability in real-world environments. In this work, we identify and analyze the unique challenges that affect the generalization of range-view LiDAR segmentation in severe weather. To address these challenges, we propose a modular and lightweight framework that enhances robustness without altering the core architecture of existing models. Our method reformulates the initial stem block of standard range-view networks into two branches to process geometric attributes and reflectance intensity separately. Specifically, a Geometric Abnormality Suppression (GAS) module reduces the influence of weather-induced spatial noise, and a Reflectance Distortion Calibration (RDC) module corrects reflectance distortions through memory-guided adaptive instance normalization. The processed features are then fused and passed to the original segmentation pipeline. Extensive experiments on different benchmarks and baseline models demonstrate that our approach significantly improves generalization to adverse weather with minimal inference overhead, offering a practical and effective solution for real-world LiDAR segmentation.}


\maketitle

\input{sec/1_intro}

\input{sec/2_relatedwork}

\input{sec/3_method}

\input{sec/4_exp}

\input{sec/5_conclusion}

\bibliography{sn-bib}

\end{document}

%% file: sec/1_intro.tex
\section{Introduction}
LiDAR segmentation has become an essential component in scene understanding, as it provides complementary 3D geometric information that traditional 2D visual data typically lacks. By assigning semantic labels (e.g., vehicles, pedestrians, road surfaces) to individual LiDAR points, semantic segmentation helps to enhance the analysis of scene geometry and contents. Recent advancements in deep learning, coupled with the availability of specialized point-cloud semantic segmentation datasets, have driven notable progress in this area~\cite{qi2017pointnet,hu2021learning,kong2023rethinking,wu2022point,wu2024point,choy20194d,zhou2020cylinder3d,yang2023swin3d,choe2022pointmixer,sun2024efficient,behley2019semantickitti,caesar2020nuscenes}. 
Despite these advancements, a critical limitation remains: most existing models are tailored for clear weather conditions, and often experience significant performance drops when exposed to adverse environments such as rain, snow, or fog~\cite{hahner2021fog,bijelic2020seeing,xiao20233d,hahner2022lidar}. This vulnerability highlights the urgent need for more robust semantic segmentation methods that can maintain accuracy and reliability across diverse and unpredictable weather conditions, which is a critical requirement for real-world deployment.

\begin{figure}
    \centering
    \includegraphics[width=\linewidth]{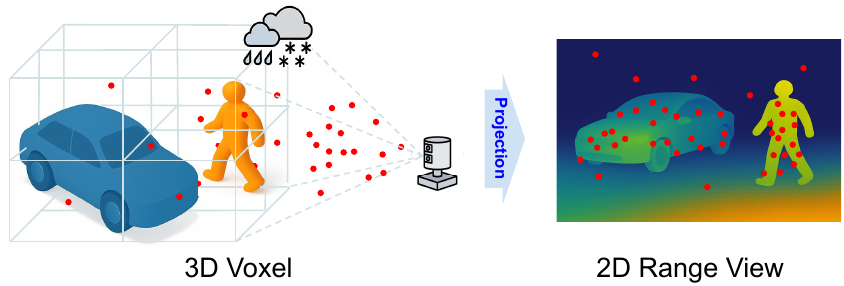}
    \caption{A simplified illustration of weather-induced noise (red dots) in LiDAR data. In 3D voxel space, noise is distributed across 3D grids and tends to be diluted among many valid ones. In contrast, range-view projection compresses 3D points into dense 2D pixels, where noise becomes more concentrated and prominent, making it more likely to interfere with the subsequent representation.  }
    \label{fig:teaser}
    \vspace{-0.6cm}
\end{figure}

To improve generalization under adverse weather, prior efforts have explored physics-based simulation~\cite{zhao2024unimix,hahner2022lidar, hahner2021fog, yang2023realistic} and task-agnostic data augmentation~\cite{xiao20233d,park2024rethinking,he2024domain}. While these strategies offer promising results, physics-based simulations often fall short in capturing the diverse scope and severity of real-world weather conditions, and augmentation-based approaches may incur extra training overhead and irrelevant variations. Moreover, these methods primarily focus on voxel-based segmentation model~\cite{xiao20233d,park2024rethinking,zhao2024unimix,park2025no}, overlooking the more efficient range-view-based alternatives. In comparison, range-view methods significantly reduce computational costs and simplify deployment by leveraging well-optimized and widely supported standard operations like 2D convolution, making them especially suitable for latency-sensitive and resource-constrained applications.

Although range-view-based methods offer clear efficiency advantages for real-world applications, enhancing their robustness and generalization under adverse-weather remains challenging and largely underexplored. A seemingly straightforward approach is to apply existing simulation-based or augmentation-based strategies~\cite{xiao20233d,park2024rethinking}. However, these voxel-oriented techniques may overlook the distinctive difficulties inherent to range-view representations. Voxel-based methods partition 3D space into regular grids, preserving spatial relationships and dispersing weather-induced noise across the volume grid. This structure naturally mitigates the impact of spurious points caused by rain, snow, or fog. In contrast, as illustrated in Fig.~\ref{fig:teaser}, range-view methods project 3D points onto a 2D plane by mapping laser beams based on their angular coordinates. In doing so, each pixel may represent one or more points, often with a preference for those nearest to the sensor or having the strongest reflectance. When handling adverse conditions, noisy points can dominate the representation and the reflectance intensity distortions can be amplified, making range-view methods more susceptible to weather variations. As a result, directly applying previous generalization strategies to range-view models may lead to suboptimal performance.

To address the challenges of adverse-weather while preserving the efficiency advantages of range-view methods, we propose a novel framework that enhances the generalization of range-view-based LiDAR semantic segmentation models, without relying on complex weather simulations or exhaustive data augmentation. Specifically, we treat LiDAR points corrupted by harsh weather as outliers and introduce a Geometric Abnormality Suppression (GAS)  module that dynamically identifies and down-weights their impact, preventing artificially amplified noise from overshadowing valid points.
Moreover, simply suppressing noisy points may lead to information loss. To compensate for this, we introduce a Reflectance Distortion Calibration (RDC)  module that leverages reflectivity as complementary information. RDC employs memory-guided adaptive instance normalization to adjust feature statistics based on stored historical domain knowledge, maintaining consistency across environmental conditions while preserving discriminative features.
Both modules are lightweight and model-agnostic, incurring minimal computational overhead. 
As a result, our method significantly improves the generalization of range-view-based LiDAR semantic segmentation towards adverse weather, without sacrificing source-domain accuracy or inference efficiency.
Our contributions can be summarized as follows: \begin{itemize} 
    \item To the best of our knowledge, we are the first to investigate generalized LiDAR semantic segmentation for range-view-based models under adverse-weather conditions. 
    \item We propose a novel framework comprising the Geometric Abnormality Suppression module and the Reflectance Distortion Calibration module to address the challenges of noisy points and reflectance distortions in range-view representations. 
    \item We demonstrate the effectiveness of our approach by applying it to different range-view-based segmentation models. Extensive experiments show that, when trained solely on clean-weather data, our method significantly improves the generalization performance across diverse weather conditions, while introducing negligible computational overhead.
\end{itemize}

%% file: sec/2_relatedwork.tex
\section{Related Work}
Existing point cloud semantic segmentation methods can broadly be categorized into three types: point-based methods, voxel-based methods and projection-based methods. 
Point-based methods directly take 3D point clouds as input and leverage Multi-Layer Perceptrons (MLPs)~\cite{qi2017pointnet,qi2017pointnet++} or point convolutions~\cite{thomas2019kpconv,wu2019pointconv,liu2020closer} to perform point-wise feature extraction. Graph Neural Networks (GNNs)~\cite{wang2019dynamic,landrieu2018large,du2022novel} and Transformer~\cite{zhao2021point,wu2022point,wu2024point,lai2022stratified,robert2023efficient} are also widely used to model relationships between points, thereby enhancing feature representation and segmentation accuracy. However, these methods often demand substantial computational resources and memory, limiting their scalability for large-scale point cloud processing in real-world applications. 
Voxel-based methods~\cite{wu20153d,choy20194d,lai2023spherical,han2020occuseg} convert irregular point clouds into structured 3D grids and utilize voxels instead of primitive points, allowing the application of 3D convolutional operations. However, due to the inherent sparsity of point clouds, traditional dense 3D convolutions often result in significant computational redundancy and memory overhead. To address these challenges, researchers have introduced 3D sparse convolutions~\cite{choy20194d,graham20183d} as a more efficient alternative, effectively reducing resource consumption while preserving performance. 
Projection-based methods transform 3D point clouds into a 2D plane, primarily using the Range View representation. These methods~\cite{cortinhal2020salsanext,milioto2019rangenet++,li2025rapid,ando2023rangevit,sun2024efficient,cheng2022cenet,kong2023rethinking} typically employ spherical projection techniques to transform discrete and sparse 3D points into a regular and dense 2D image, which can then be processed using 2D Convolutional Neural Network. Although this conversion process results in a certain degree of information loss, the range-view-based approach achieves the fastest inference speed with the lowest memory consumption.

While existing LiDAR-based perception methods exhibit strong performance under normal weather conditions, their reliability in adverse weather remains a critical challenge~\cite{yan2024benchmarking, kong2023robo3d, xiao20233d}. Due to the lack of training data under extreme conditions, the performance of the previous method has seriously declined in scenarios such as fog, rain and snow. To bridge this data gap, simulation-based solutions have been proposed by leveraging physical models to synthesize degraded data. For instance, Dreissig~\etal~\cite{yang2023realistic} and Hahner~\etal~\cite{hahner2022lidar, hahner2021fog} explore methods for simulating rain, fog, and snow effects from normal-condition data, while Unimix~\etal~\cite{zhao2024unimix} incorporates synthetic artifacts during training. However, these methods often oversimplify the complexity of real-world weather, limiting their ability to model domain shifts accurately.
Recent work increasingly explores domain-agnostic augmentations to improve generalization. PointDR~\cite{xiao20233d} and RDA~\cite{park2024rethinking} focus on geometric augmentation with the former employing randomized geometry styles and the latter utilizing selective jittering and learnable point dropping in geometric space, while DGUIL~\cite{he2024domain} addresses domain shifts by modeling point uncertainty through uncertain domain distributions in feature space. 
Some methods also adopt general learning frameworks, such as teacher-student models~\cite{zhao2024unimix}, to improve robustness.
However, it is important to note that these approaches have not specifically addressed the generalization ability of range-view-based LiDAR segmentation models under adverse weather. As a result, their performance in extreme conditions remains suboptimal, underscoring the need for dedicated solutions tailored to the unique challenges of range-view representations.

%% file: sec/3_method.tex
\begin{figure*}[t]
  \centering
  \includegraphics[width=0.95\linewidth]{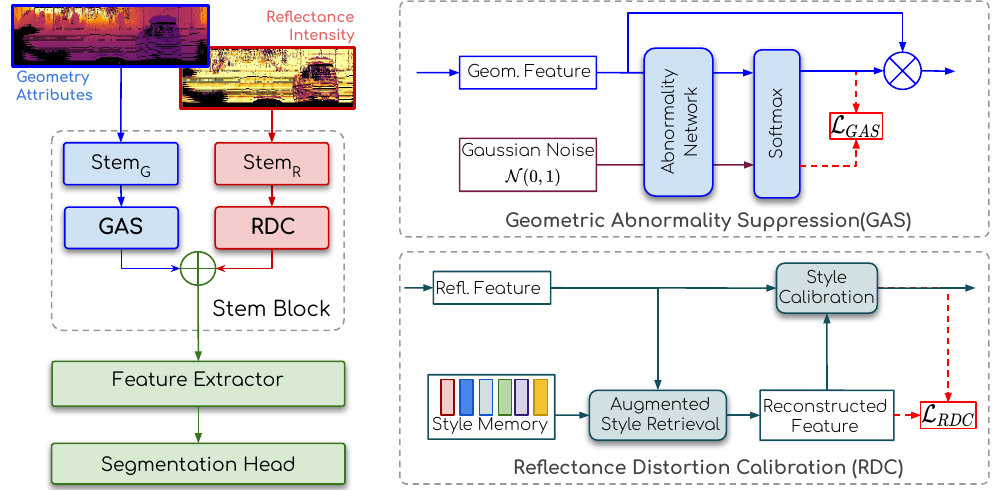}
  \caption{ Overview of the Proposed Method. The framework operates only at the initial stage of existing models to process geometric attributes and reflectance intensity separately, thereby maintaining compatibility with mainstream architectures. The geometric branch is processed by the Geometric Abnormality Suppression (GAS) module to reduce noise caused by adverse-weather conditions, while the reflectance branch is handled by the Reflectance Distortion Calibration (RDC) module to correct domain shifts in reflectance intensity. The processed features are then fused and passed through the original feature extractor and segmentation head, ensuring enhanced robustness while keeping the existing models’ efficiency.} 
  \label{fig:overview}
\end{figure*}

\section{Methodology}

In this section, we introduce the proposed framework for enhancing the generalization of range-view-based LiDAR semantic segmentation under adverse-weather conditions. As depicted in the left part of Fig.~\ref{fig:overview}, the key idea is to process geometric attributes and reflectance intensity separately within stem block ( the initial set of shallow layers for processing raw input data) of existing models, while leaving the feature extraction stages and segmentation head of typical range-view architectures unchanged. This modular design ensures compatibility with mainstream models~\cite{cortinhal2020salsanext,ando2023rangevit,cheng2022cenet,milioto2019rangenet++}, and allows for easy integration with minimal overhead.

In particular, the range-view representation of input LiDAR is split into two parallel branches: one for geometric attributes and the other for reflectance intensity. The geometric branch passes through the Geometric Abnormality Suppression (GAS) module, which identifies and down-weights noisy or abnormal points that typically arise under adverse-weather conditions. In parallel, the reflectance branch is processed by the Reflectance Distortion Calibration (RDC) module, which mitigates domain shifts in reflectance intensity by leveraging a memory-guided adaptive instance normalization strategy. These processed features are then fused and passed through the original feature extractor and segmentation head, allowing the network to benefit from enhanced robustness without compromising its original design or efficiency.

\subsection{Preliminaries}
\label{sec:method_fomulation}
We consider the task of domain-generalized LiDAR semantic segmentation. The objective is to train a segmentation model using only source domain data collected under clear-weather conditions, such that it generalizes effectively to unseen target domains characterized by adverse weather (e.g., rain, snow, fog), while maintaining high accuracy on the source domain and incurring minimal additional inference latency.
Formally, we are given a labeled source domain dataset $\mathcal{S} = \{(P^{s}_{i}, L^{s}_{i})\}_{i=1}^{N^{s}}$, where $N^s$ denotes the number of LiDAR scans. Each sample $(P^{s}_{i}, L^{s}_{i})$ consists of a point cloud $P^{s}_{i}$ containing $n$ points, and corresponding point-wise semantic labels $L^{s}_{i}$. The point cloud is defined as $P^{s}_{i} = \{(x^{s}_{ij}, y^{s}_{ij}, z^{s}_{ij}, r^{s}_{ij})\}^{n}_{j=1}$, where $(x, y, z)$ denotes the 3D coordinates and $r$ represents the reflectance intensity of each point. The goal is to learn a semantic segmentation model $f: P \rightarrow L$ based solely on the source domain data $\mathcal{S}$, such that it performs well on an unlabeled target domain $\mathcal{T} = \{P^{t}_{i}\}_{i=1}^{N^{t}}$, which remains entirely unseen during training.

To enable efficient processing of LiDAR point clouds using standard 2D deep learning operations, range-view-based methods project irregular 3D point sets onto a dense 2D representation. This projection maps each LiDAR point to a pixel on a spherical image, where horizontal and vertical axes correspond to the azimuth and elevation angles of the sensor field of view.
Given a point $p_i = (x_i, y_i, z_i, r_i)$ with Cartesian coordinates and reflectance intensity, its corresponding pixel location $(u_i, v_i)$ in the range image is computed as:
\begin{equation}
\begin{aligned}
    u_i &= \frac{1}{2}\left[1 - \frac{\arctan(x_i, y_i)}{\pi} \right] W \\
    v_i &= \left[1 - \frac{\arcsin(z_i / d_i) + \phi^{\text{down}}}{\xi} \right] H,
\end{aligned}
\end{equation}
where $d_i = \sqrt{x_i^2 + y_i^2 + z_i^2}$ is the range (depth) of the point, $\phi^{\text{up}}$ and $\phi^{\text{down}}$ are the upward and downward vertical field-of-view limits of the LiDAR, and $\xi = |\phi^{\text{up}}| + |\phi^{\text{down}}|$ is the total vertical FOV. The image dimensions are denoted by $H$ (height) and $W$ (width).
Due to the radial nature of the projection, multiple 3D points may map to the same 2D pixel. In such cases, a common practice is to retain the point with the closest distance or strongest intensity.

Building on such projection mechanisms, while efficient, range-view representations can exhibit unique vulnerabilities under adverse-weather conditions. For geometric attributes, noise points caused by rain, snow, or fog may occlude or overshadow valid neighboring points, resulting in distorted spatial structures and incomplete object appearances. For reflectance intensities, the dense arrangement of pixels in the range view magnifies weather-induced fluctuations, causing amplified distortions and exacerbating domain shifts of reflectance.  These combined effects can significantly degrade segmentation performance, highlighting the need for specialized strategies for generalized range-view understanding.

\subsection{Geometric Abnormality Suppression}
\label{sec:method_gas}
Adverse weather conditions such as rain, snow, and fog introduce significant noise into LiDAR point clouds, largely due to the scattering and reflection of laser pulses by airborne particles. Operating in the near-infrared (NIR) spectrum, LiDAR systems suffer from poor transmission through water-based media~\cite{hahner2021fog,hahner2022lidar}, leading to backscattered signals and the generation of spurious returns. For example, raindrops can produce isolated noise points suspended in the air, snowflakes yield dense and intense noise due to their larger size and irregular shapes, and fog droplets cause strong Mie scattering~\cite{drake1985mie}, resulting in global energy attenuation and reduced detection range. Rain accumulation on surfaces may also create multipath reflections, producing farther but invalid returns. These effects degrade geometric fidelity.

To suppress these noisy geometric features, we propose the Geometric Abnormality Suppression (GAS) module. GAS aims to dynamically identify and down-weight features likely originating from weather-induced artifacts. It employs a lightweight binary classifier to predict the likelihood that each pixel corresponds to a valid point, based on geometric cues. However, since weather noise is scarce in source-domain training data, it is difficult to directly learn a discriminator in a supervised way. 
To address this, we leverage the observation that geometric features extracted from clean-weather scenes exhibit consistent spatial patterns, while weather-induced noisy points tend to deviate from these norms. Rather than relying on explicit supervision, we construct a self-supervised learning task that contrasts structured geometric signals against randomized noise. As illustrated in the right-top part in Fig~\ref{fig:overview}, we treat the original geometry-only features processed by initial layers $Stem_G(\cdot)$ as positive examples, and generate challenging negatives by sampling from a standard Gaussian distribution. This allows the model to implicitly learn a boundary between structured and anomalous geometric feature patterns, helping the classifier to identify weather-corrupted points without exposure to real-world noise during training.

Formally, begin by applying the initial block $Stem_G(\cdot)$ to extract dense geometric feature  $F_{\text{geo}} \in \mathbb{R}^{C \times H \times W}$ from geometric attributes (e.g depth and XYZ coordinates) represented in the range-view form. Then, positive samples are generated as:
\begin{equation}
    F^+ = F_{\text{geo}} + \gamma \cdot \epsilon, \quad \epsilon \sim \mathcal{N}(0, \mathbf{I}).
\end{equation}
where $\mathbf{I}$ is a $D \times D$ identity matrix, $\epsilon$ is sampled from a standard Gaussian, and $\gamma$ is a small coefficient that we empirically set to be 0.02. By adding the slight perturbations to the feature of source domain features to be positive samples, we encourage the classifier to learn a compact representation of clean geometric structures tolerant to minor variations. 

In the meantime, negative samples are directly sampled from the normal distribution : 
\begin{equation}
    f^- \sim \mathcal{N}(\mu_n, \delta_n^2\cdot\mathbf{I}).
\end{equation}
where $\mu_n$ and $\delta_n$ are  parameters of the Gaussian, and $\mathbf{I}$ is a $D \times D$ identity matrix. In practice, we utilize the standard Gaussian by setting $\mu_n=0$ and $\delta_n=1$. The negative samples are sampled multiple times to form the negative feature map $F^-\in \mathbb{R}^{ C \times H \times W}$

Then, we concatenate the two types of features along the batchsize dimension to form a training batch:
\begin{equation}
    F_{\text{train}} = \text{Concat}(F^+, F^-) \in \mathbb{R}^{2 \times C \times H \times W}.
\end{equation}

And an Abnormality Network $f_\theta(\cdot)$ formulated by several $1\times1$ convolutional blocks followed by a softmax layer is applied to distinguish between them:
\begin{equation}
    [\mathbf{S}^+,\mathbf{S}^-] = f_\theta(F_{\text{concat}}) 
\end{equation}
where $\mathbf{S}^+\in \mathbb{R}^{2 \times H \times W}$ and $\mathbf{S}^-\in \mathbb{R}^{2 \times H \times W}$ denote the predicted logits for positive and negative samples, respectively. And $f_\theta(\cdot)$ is optimized via the classification loss:
\begin{equation}
    \mathcal{L}_{\text{GAS}} = \mathcal{L}_{\text{CE}}(\mathbf{S}^+_M, \mathbf{1}) + \mathcal{L}_{\text{CE}}(\mathbf{S}^-_M, \mathbf{0})
\end{equation}
where $\mathcal{L}_{\text{CE}}$ is the cross-entropy loss.

At inference time, the abnormality classifier is applied to the geometric feature map, and its predicted probabilities $\mathbf{W}^+ \in \mathbb{R}^{ H \times W}$ of being normal points are utilized to weight the corresponding original geometric features:
\begin{equation}
    \hat{F}_{geo} = \mathbf{W}^+ \odot F_{geo}
\end{equation}
which is later processed by subsequent modules of segmentation models.

In doing so, the Geometric Abnormality Suppression (GAS) module effectively differentiates between structured geometric features and anomalous patterns without exposing to adverse-weather domains. This approach is theoretically supported by the manifold hypothesis suggesting that high-dimensional data lie on low-dimensional manifolds within the data space, and normal data points are expected to reside on these manifolds, while anomalies deviate from them~\cite{hojjati2024self,ruff2021unifying}. The GAS module's self-supervised approach implicitly encourages the learning of representations that map normal geometric structures onto a compact manifold. Consequently, points that do not conform to this learned manifold, such as those affected by adverse-weather conditions, can be identified as anomalies.

\subsection{Reflectance Distortion Calibration}
\label{sec:method_rdc}
While the GAS module can attenuate the influence of weather-induced noise by down-weighting unreliable geometric features, this suppression inevitably leads to the loss of potentially informative points, particularly at object surfaces or in regions with sparse returns. To compensate for this loss and reinforce semantic consistency, we turn to reflectance intensity, which serves as a complementary cue that can enhance scene understanding under challenging conditions.

However, incorporating reflectance in a domain-generalized framework introduces unique challenges, particularly in range-view representations. Unlike geometric attributes, reflectance is highly sensitive to external factors such as surface materials, angles of incidence, and most notably, weather-induced attenuation. In range-view projections, this sensitivity is further compounded by densely packed pixels, where even slight variations in reflectance tend to propagate across neighboring pixels. As a result, localized anomalies accumulate into systematic distortions that span large regions of the reflectance map. Consequently, these aggregated discrepancies lead to a significant domain shift between clear weather and adverse-weather conditions. To address this, we propose a Reflectance Distortion Calibration (RDC) module that adaptively normalizes reflectance features across different environmental conditions while preserving their semantic utility. The key idea is to leverage memory-guided normalization, informed by previously observed clean-weather statistics, to calibrate distorted features during inference, thereby achieving robust performance.

Inspired by Adaptive Instance Normalization~\cite{huang2017arbitrary},  where aligning the mean and variance of an input (content) feature map to those of a target (style) feature map can “transfer” the style of the latter onto the former, we propose a novel memory-guided approach as illustrated in right-bottom part of Fig.~\ref{fig:overview} to adjust reflectance distributions under diverse weather conditions. Treating the reflectance features from adverse environments as distinct styles, our method adaptively aligns them with source-domain style statistics. This alignment is achieved through retrieving and apply the most relevant source-domain style vectors from a learnable memory bank, enabling a transformation of input reflectance features into a source-domain style. 

Formally, let $F_{\text{ref}} \in \mathbb{R}^{C \times H \times W}$ be the reflectance feature map extracted by applying the reflectance-related stem layers $Stem_R(\cdot)$ to the input intensity. Our goal is to adaptively apply source-domain style statistics that are semantically aligned with $F_{\text{ref}}$, thereby adjusting its reflectance feature distribution to match the source domain. Concretely,
\begin{align}
    \hat{F}_{\text{ref}}=\sigma_{src}(F_{\text{ref}})(\frac{F_{\text{ref}}-\mu_{\text{ref}}}{\sigma_{\text{ref}}})+\mu_{src} (F_{\text{ref}})
    \label{eq:transfer}
\end{align}
where $\mu_{\text{ref}}$ and $\sigma_{\text{ref}}$ denote the channel-wise statistics of $F_{\text{ref}}$.  Meanwhile, $\mu_{\mathrm{src}}(F_{\text{ref}})$ and $\sigma_{\mathrm{src}}(F_{\text{ref}})$ represent the input-correlated style statistics of the source domain, which are adaptively estimated to achieve robust alignment under varying conditions. To this end, we introduce a parametric Source Style Memory Bank $\mathcal{M} \in \mathbb{R}^{T \times C}$, where $T$ is the number of style vectors and $C$ is the feature dimension. This memory bank stores style vectors learned on the source domain. 

\vspace{0.2em}
\noindent\textbf{Augmented Style Retrieval}.\quad To adaptively align reflectance features with source-domain statistics, we employ a memory-based retrieval mechanism, augmented by random perturbations for broader generalization. 

For each  feature  $f_{\text{i}} \in \mathbb{R}^{C}$ from an input feature map $F_{\text{ref}}\in \mathbb{R}^{C \times H \times W}$, we compute its cosine similarity $s_{\text{ij}} = \frac{m_\text{j} \cdot f_\text{i}}{\|m_\text{j}\|\|f_\text{i}\|}$ to source style vectors in the memory bank $\mathcal{M}\in \mathbb{R}^{T \times C}$, resulting in a similarity matrix $S \in \mathbb{R}^{T \times H \times W}$. Applying a softmax along the $T$ dimension yields the attention map $A=\text{Softmax}(S) \in \mathbb{R}^{T \times H \times W}$.
We then compute the retrieved style map $V$ by,
\begin{equation}
    V \;=\;\mathcal{M}^{T} \cdot A, \quad V \in \mathbb{R}^{C \times H \times W}.
\end{equation}

This spatially varying mapping transfers relevant style information from the source domain into $V$, producing locally aligned style vectors for each pixel location. Subsequently, we obtain the input-related source statistics from $V$ in a channel-wise manner,
\begin{equation}
    \begin{aligned}
    \mu_{\mathrm{src},c}(F_{\text{ref}}) &= \frac{1}{HW}\,\sum_{i=1}^{H}\sum_{w=1}^{W} V_{chw},\\
    \sigma_{\mathrm{src},c}(F_{\text{ref}})&=\sqrt{ \frac{1}{HW}\,\sum_{h=1}^{H}\sum_{w=1}^{W}(V_{chw}-\mu_{\mathrm{src},c}(F_{\text{ref}}))^2}
    \end{aligned}
\end{equation}

Since the training data contains only clear-weather samples, the style memory bank may overfit by reconstructing the exact source distribution rather than learning generalized representations. To mitigate this, instead of directly utilizing the input feature for style retrieval during training, we further augment $F_{\text{ref}}$ by randomly perturbing its channel-wise mean and standard deviation,

\begin{align}
    {F}_\text{ref}^\text{aug}=\sigma_{\text{aug}}(\frac{F_{\text{ref}}-\mu_{\text{ref}}}{\sigma_{\text{ref}}})+\mu_{\text{aug}}
    \label{eq:transfer}
\end{align}
where $\mu_{\text{ref}}$ and $\sigma_{\text{ref}}$ denote the statistics of input feature map $F_{\text{ref}}$. Meanwhile, $\mu_{\text{aug}}=(\alpha+0.5) \cdot \mu_{\text{ref}}$ and $\sigma_{\text{aug}}=(\beta+0.5) \cdot \sigma_{\text{ref}}$ are the augmented statistics, 
with $\alpha$ and $\beta$ are sampled from a uniform distribution $\mathcal{U}(0,1)\in\mathbb{R}^{C}$.


\vspace{0.2em}
\noindent\textbf{Style Calibration}.\quad 
After retrieving the source-domain statistics $\mu_{\mathrm{src}}(F_{\text{ref}}^\text{aug})$ and $\sigma_{\mathrm{src}}(F_{\text{ref}}^\text{aug})$ from the augmented style retrieval step, we perform a final calibration of the original reflectance features. Specifically, we replace the channel-wise mean and variance of $F_{\text{ref}}$ with the learned source statistics, transforming $F_{\text{ref}}$ into a clean-weather style:
\begin{equation}
\label{eq:calibration}
    \hat{F}^\text{aug}_{\text{ref}} \;=\; \sigma_{\mathrm{src}}(F^\text{aug}_{\text{ref}}) \left(\frac{F_{\text{ref}} - \mu_{\text{ref}}}{\sigma_{\text{ref}}}\right) \;+\; \mu_{\mathrm{src}}(F^\text{aug}_{\text{ref}}),
\end{equation}
where $\mu_{\text{ref}}$ and $\sigma_{\text{ref}}$ denote the channel-wise mean and standard deviation of $F_{\text{ref}}$.
By replacing the native reflectance statistics with source-aligned counterparts, $\hat{F}^\text{aug}_{\text{ref}}$ becomes more robust to distribution shifts induced by adverse weather. Notably, this process preserves the semantic structure of the underlying features, relying only on channel-wise normalization rather than altering spatial arrangements. Consequently, the model gains stronger resilience to reflectance distortion without sacrificing critical object-level cues.

\input{table/sk2stf}

To ensure effective style adaptation while preserving semantic fidelity, we introduce two complementary loss functions. First, the \emph{Semantic Consistency Loss} enforces the stylized feature $\hat{F}^\text{aug}_{\text{ref}}$ to retain the original semantic content of $F_{\text{ref}}$:
\begin{equation}
    \mathcal{L}_{SC} \;=\; \bigl\|\hat{F}^\text{aug}_{\text{ref}} - F_{\text{ref}}\bigr\|.
\end{equation}
In addition, we employ a \emph{Style Alignment Loss} to ensure that the transferred features align with the source-domain statistics:
\begin{equation}
    \begin{split}
        \mathcal{L}_{SA} \;=\; \bigl\|\mu_{\mathrm{src}}\bigl(F^\text{aug}_{\text{ref}}\bigr) - \mu\bigl(F_{\text{ref}}\bigr)\bigr\|
        ~~~\\ \;+\;
        \bigl\|\sigma_{\mathrm{src}}\bigl(F^\text{aug}_{\text{ref}}\bigr) - \sigma\bigl(F_{\text{ref}}\bigr)\bigr\|.
    \end{split}
\end{equation}
Finally, the overall loss for the RDC module combines these terms:
\begin{equation}
    \mathcal{L}_{RDC} 
    \;=\;
    \mathcal{L}_{sc} 
    \;+\;
    \mathcal{L}_{sa}.
\end{equation}

By jointly optimizing these constraints on the source domain, the style memory bank learns a disentangled and transferable representation of source-domain styles, while preserving the model's discriminative capacity across varying weather conditions.

\subsection{Training}
\label{sec:training}

In the feature preprocessing stage, we first extract and adjust the geometric and reflectance features independently using their respective modules. The processed features, denoted as $\hat{F}_{geo}$ for geometry and $\hat{F}^\text{aug}_{\text{ref}}$ for reflectance (obtained with augmentation), are then fused via point-by-point summation to produce a unified representation. This fused feature map is subsequently passed to subsequent layers for semantic segmentation.

During training, the augmented reflectance features $\hat{F}^\text{aug}_{\text{ref}}$ are used to expose the model to a broader range of reflectance styles, thereby improving generalization under adverse-weather conditions. Note that such augmentation is applied only during training; at test time, the model operates on the unaugmented reflectance features. The overall training loss on the source data is defined as:
\begin{equation}
    \mathcal{L} \;=\; \mathcal{L}_{GAS} \;+\; \mathcal{L}_{RDC} \;+\; \mathcal{L}_{SEG},
\end{equation}
where $\mathcal{L}_{GAS}$ is the loss associated with the Geometric Abnormality Suppression module, $\mathcal{L}_{RDC}$ is the loss for the Reflectance Distortion Calibration module, and $\mathcal{L}_{SEG}$ denotes the segmentation task loss (e.g., cross-entropy loss).

%% file: table/sk2stf.tex
\renewcommand\arraystretch{1.2}
\setlength{\tabcolsep}{0.95mm}{
\begin{table*}[t]
    \centering
    \begin{footnotesize}
    \resizebox{1\textwidth}{!} {   
    \begin{tabular}{l||ccccccccccccccccccc|cccc|c||c}
        \toprule
        Method  & \rotatebox{90}{car} & \rotatebox{90}{bi.cle} & \rotatebox{90}{mt.cle} & \rotatebox{90}{truck} & \rotatebox{90}{oth-v.} & \rotatebox{90}{pers.} & \rotatebox{90}{bi.clst} & \rotatebox{90}{mt.clst} & \rotatebox{90}{road} & \rotatebox{90}{parki.} & \rotatebox{90}{sidew.} & \rotatebox{90}{oth-g.} & \rotatebox{90}{build.} & \rotatebox{90}{fence} & \rotatebox{90}{veget.} & \rotatebox{90}{trunk} & \rotatebox{90}{terra.} & \rotatebox{90}{pole} & \rotatebox{90}{traf.} & \rotatebox{90}{D-fog} & \rotatebox{90}{L-fog} & \rotatebox{90}{Rain} & \rotatebox{90}{Snow} & Avg & Source\\
        \hline
        CENet~\cite{cheng2022cenet} &  18.6 & 0.0 & 0.0 & 8.5 & 0.0 & 0.0 & 0.0 & 0.0 & 29.4 & 1.7 & 11.7 & 2.1 & 20.5 & 14.0  & 10.1 & 2.1 & 18.0 & 8.1 & 3.1 & 15.7 & 8.4 & 7.9 & 3.2 &  7.8 & \textbf{60.9} \\
        \quad w/ PDR~\cite{xiao20233d} & 6.5 & 0.3 & 0.0 & 0.1 & 2.1 & 0.0 & 0.0 & 0.0 & 11.5  & 0.5 & 3.9 & 0.2 & 23.7  & 5.5 & 13.8 & 1.3 & 19.9 & 17.1 & 1.2 & 8.9 & 4.8 & 2.4 & 3.0 & 5.7 & 60.1 \\
        \quad w/ RDA~\cite{park2024rethinking} & 18.0 & \textbf{5.5} & 0.0 & 12.9 & 3.0 & 1.3 & \textbf{27.4} & \textbf{29.3} & 48.0 & 8.5 & 21.5 & 1.7 & 33.5 & 23.1 & \textbf{50.0} & 13.5 & 38.1 & 11.8 & 0.0 & 28.9 & 17.8 & 13.7  & 10.9 & 18.3 & 58.1 \\
        \quad w/ GRC~\cite{yang2025towards} & \textbf{48.3} & 0.4 & \textbf{0.6} & 0.0 & \textbf{14.5} & 28.7 & 0.0 & 0.0 & \textbf{65.9} & 11.9 & \textbf{35.2} & 0.3 & 52.8 & 20.6 & 32.2 & 20.6 & 38.4 & 17.5 & 19.3 & 19.5 & 21.6 & 22.4 & \textbf{22.6} & 21.4 & 59.5 \\
        \quad  w/ Ours & 39.0 & 0.0 & 0.0 & \textbf{19.8} & 8.7 & \textbf{42.1} & 12.1 & 8.4 & 44.5 & \textbf{12.5} & 23.0 & \textbf{4.1} & \textbf{62.8} & \textbf{40.8} & 46.9 & \textbf{32.9} & \textbf{40.0} & \textbf{23.9} & \textbf{24.5} & \textbf{30.0} & \textbf{25.5} & \textbf{23.7} & 15.8 &  \textbf{25.6} & 60.5 \\
        \hline
        RangeNet$^{\text{++}}$~\cite{milioto2019rangenet++} & 15.9 & 0.1 & 0.0 & 0.2 & 1.8 & 0.0 & 0.0 & 0.0 & 8.7 & 0.2 & 5.6 & 2.5 & 29.0 & 8.7 & 11.4 & 4.9 & 29.8 & 11.6 & 7.6 & 13.0 & 7.5 & 7.4 & 4.5 & 7.3 & 46.5 \\
       \quad   w/ PDR~\cite{xiao20233d} & 7.3 & 0.0 & 0.0 & 0.0 &  0.0 & 0.0  & 0.0& 0.0 & 11.8 & 0.5 & 5.4 & 2.2 & 20.4 & 5.9 & 10.5 & 0.6 & 15.3 & 2.4 & 7.2 &9.0 & 3.0  & 2.3  &  2.2 & 4.7 & 46.5\\
        \quad w/ RDA~\cite{park2024rethinking} & 34.0 & 0.0 & 0.0 & 2.7 & 2.2 & 0.7 & 0.0 & 0.3 & 56.8 & 2.3  & 22.4 & 3.0 & 53.1 & 20.9 & 43.7 & 11.8 & 40.9 & 8.9 & 0.0 & 22.9 & 18.47 & 17.7 & 13.2 & 16.0 & 45.9\\
        \quad w/ GRC~\cite{yang2025towards} & 23.5 & 0.0 & 0.0 & 3.0 & 3.2 & 5.3 & 0.0 & 0.0 & 36.4 & 2.6 & 13.7 & 2.8 & 42.9 & 21.7 & 26.0 & 12.7 & 28.4 & 11.1 & 18.2 & 18.4 & 14.5 & 10.7 & 12.3 & 13.2 & 43.2 \\
        \quad  w/ Ours & \textbf{72.9} & \textbf{0.2}  &  \textbf{0.1} & \textbf{14.0} & \textbf{14.2} & \textbf{22.2} & 0.0 & \textbf{1.0} & \textbf{68.0} & \textbf{10.1} & \textbf{35.3} & \textbf{4.7} & \textbf{66.8} & \textbf{32.8} & \textbf{49.6} & \textbf{21.6} & \textbf{44.3 }& \textbf{24.2} & \textbf{18.9} & \textbf{26.0} & \textbf{25.9} & \textbf{26.8} & \textbf{28.1} & \textbf{26.4} & \textbf{46.9}\\
        \hline
        SalsaNext~\cite{cortinhal2020salsanext} & 14.6 & 0.0 & 0.0 & 8.3 & 2.4 & 0.0 & 0.0 & 0.0 & 28.3 & 5.4 & 10.7 & 0.6 & 14.2 & 10.2 & 14.1 & 2.7 & 22.0 & 15.8 & 0.9 & 14.3 & 8.1 & 7.0  & 3.9 & 7.9 & 59.7\\
        \quad  w/ PDR~\cite{xiao20233d} & 14.8 & 0.0 & 0.0 & 2.3 & 1.8 & 0.0 & 0.0 & 0.8 & 17.9 & 1.0 & 4.2 & 1.8 & 12.8 & 10.6 & 14.2 & 5.2 & 21.8 & 7.5 & 1.0 & 10.7 & 5.1 & 3.2 & 1.9 & 6.2 & 58.5 \\
        \quad w/ RDA~\cite{park2024rethinking} & 21.5 & 0.1 & 0.0 & \textbf{40.3} & 9.7 & 0.0 & 0.0 & 0.0 & 47.9 & 0.2 & 17.2  & 2.4 & 38.3 & 30.2 & 42.2 & 11.9 & 42.4 & 28.6 & 2.1 & 25.5 & 22.6 & 15.9 & 10.6 &17.6 & 54.6\\
        \quad w/ GRC~\cite{yang2025towards} & 54.1 & 0.0 & 0.0 & 2.5 & \textbf{20.3} & 15.7 & 0.0 & \textbf{54.2} & \textbf{53.5} & 10.4 & 22.4 & 1.3 & 44.4 & 28.1 & 40.9 & 13.6 & 38.0 & 9.8 & 5.7 & 23.6 & 19.9 & 21.0 & 18.1 & 21.9 & 58.4 \\
        \quad  w/ Ours &  \textbf{60.6} & \textbf{0.1} & 0.0 & 38.3 & 13.3 & \textbf{41.2} & 0.0 & 18.3 & 47.2 & \textbf{16.8} & \textbf{28.6} & \textbf{3.1} & \textbf{60.8} & \textbf{33.8} & \textbf{50.3} & \textbf{27.1} & \textbf{42.9} & \textbf{30.1} & \textbf{22.6} & \textbf{30.2}  & \textbf{27.9} & \textbf{21.5} & \textbf{21.9} & \textbf{ 28.2} &\textbf{60.2}\\
        \hline
        RangeViT~\cite{ando2023rangevit} & 12.0 & 0.2 & 0.0 & 3.4 & 0.5 & 0.0 & 0.0 & \textbf{1.3} & 43.1 & 0.6 & 15.3 & 1.2 & 33.0 & 12.7  & 26.2 & 3.0 & 25.0 & 8.0 & 8.8 & 16.5 & 10.8 & 9.3 & 6.4 & 10.2 & \textbf{59.6}\\
        \quad  w/ PDR~\cite{xiao20233d} & 17.5 & 0.0 & 0.0 & 13.0 & 2.6 & 0.0 & 1.2 & 1.1 & 34.8 & 5.9 & 13.4 & 0.7 & 34.9 & 14.1 & 23.7 & 4.5 & 29.1 & 8.6 & 15.9 & 18.4 &  13.1 & 9.1 & 6.3 & 11.6  & 58.3 \\
        \quad w/ RDA~\cite{park2024rethinking} & 60.6 & 0.0 & 0.0 & 20.8 & \textbf{18.2} & 21.5 & 0.0 & \textbf{9.1} & \textbf{62.4} & \textbf{16.7} & \textbf{36.3} & 3.5 & 66.2 & 35.3 & 50.5 & \textbf{30.6} & 37.6 & 18.9 & 0.0 & 37.0 & 29.1 & 25.7 & 24.1 & 25.7 & 53.3\\
        \quad w/ GRC~\cite{yang2025towards} & 20.6 & 0.0 & 0.0 & 10.5 & 8.4 & 0.4 & \textbf{3.1} & 0.0 & 35.2 & 3.1 & 15.3 & 1.7 & 48.6 & 25.0 & 29.9 & 12.2 & 31.5 & 11.0 & 11.4 & 22.0 & 16.5 & 13.1 & 9.0 & 14.1 & 58.2\\
        \quad  w/ Ours & \textbf{69.8} & \textbf{1.5} & \textbf{8.0} & \textbf{22.9} & 15.7 & \textbf{40.1} & 0.4 & 0.0 & 54.3 & 12.1 & 33.4 & \textbf{10.0} & \textbf{69.6} & \textbf{43.6} & \textbf{52.2} & 26.5 & \textbf{41.3} & \textbf{21.1} & \textbf{26.4}  & \textbf{29.5} &  \textbf{31.8} & \textbf{30.4} & \textbf{26.2} & \textbf{28.9} & 59.3 \\
        \bottomrule
    \end{tabular}
    }
    \caption{Generalized segmentation performance for ``SemanticKITTI$\rightarrow$SemanticSTF''. The  last column ``Source'' shows the performance on the source domain SemanticKITTI.}
    \label{tab:kitti-stf}  
  \end{footnotesize}
  \vspace{-0.1cm}
\end{table*}
}

%% file: sec/4_exp.tex
\section{Experiments}
We assess the generalization performance of our method with  four representative range-view-based models including SalsaNext~\cite{cortinhal2020salsanext},  RangeNet$^{\text{++}}$~\cite{milioto2019rangenet++}, CENet~\cite{cheng2022cenet}, and RangeViT~\cite{ando2023rangevit} on two standard benchmarks: SemanticSTF~\cite{xiao20233d}, which consists real-world adverse-weather conditions, and SemanticKITTI-C~\cite{kong2023robo3d}, which introduces simulated corrupted effects.

\subsection{Datasets and Implementation}
\label{sec:4.1}
\vspace{0.2em}
\textbf{Datasets.} \quad
\textit{SemanticKITTI}~\cite{behley2019semantickitti} is a comprehensive, large-scale dataset for semantic segmentation, derived from the KITTI Vision Benchmark~\cite{geiger2012we}. This dataset contains 64-beam LiDAR point clouds captured in Germany urban areas and  annotated with over 19 distinct semantic categories. In accordance with the official protocol, sequences 00-07 and 09-10 are designated as the training set, while sequence 08 serves as the validation set. A modified version, SemanticKITTI-C~\cite{kong2023robo3d} is created by simulating LiDAR corruption at varying severity levels.
\textit{SynLiDAR}~\cite{xiao2022transfer} represents a substantial synthetic LiDAR dataset, comprising approximately 20,000 annotated scans. It features precise geometric shapes and an extensive range of semantic classes. We use sequences 00-12 as the training set following previous works\cite{xiao20233d, park2024rethinking, zhao2024unimix, xiao2022transfer}. 
\textit{SemanticSTF}~\cite{xiao20233d} is a specialized LiDAR segmentation dataset designed to address adverse-weather conditions. It extends the realistic STF Detection Benchmark~\cite{bijelic2020seeing} by incorporating point-wise annotations for 21 semantic categories. This dataset includes four common adverse weather scenarios: dense fog, light fog, snow, and rain. Following the official protocol, we only utilize its validation set for our testing purposes. We use mean Intersection over Union (mIoU) computed on the original 3D point clouds as the evaluation metric in all experiments.

\vspace{0.2em}
\noindent\textbf{Implementation Details.} \quad
Given a range-view-based model, we only repalce its initial layers in its stem block with our propose dual-branch module as in Fig.~\ref{fig:overview}. During training, we apply standard data augmentations to the point clouds, including random dropping, rotation, flipping, and scaling. For SalsaNext, RangeNet$^{\text{++}}$, and CENet, we utilize a batch size of 6 to train the models for 50 epochs with the AdamW optimizer, a weight decay factor of 0.0001, an initial learning rate of 0.0025, and dynamically adjust the learning rate using the OneCycle strategy. For RangeViT, we follow its official configuration and train for 60 epochs. The Horizontal resolution is set as 2048 for all the experiments.

\subsection{Main Results}
\textbf{SemanticKITTI to SemanticSTF}. \quad
As shown in Table~\ref{tab:kitti-stf}, our method significantly enhances generalization performance on the SemanticSTF dataset. Compared to baseline range-view models, we observe substantial improvements in mIoU on the testing domain. These gains are largely due to our approach’s ability to suppress spatial noise and reflectance intensity distortion.
For example, compared to the original SalsaNext model, our approach shows impressive improvements under different weather conditions: +15.9 mIoU in dense fog, +19.8 mIoU in light fog, +14.5 mIoU in rain, and +18.0 mIoU in snow, with an overall average gain of +20.3 mIoU across all classes. Notably, applying our method to RangeViT results in the best performance among all range-view methods, achieving an increase  from 10.2 mIoU to 28.9 mIoU, surpassing even the voxel-based method MinkNet+PDR~\cite{xiao20233d}, which achieves 28.6 mIoU. This demonstrates that our method not only excels in improving performance on a variety of weather conditions but also competes favorably against voxel-based approaches. When evaluating the performance on source domain (SemanticKITTI), our method outperforms several baseline range-view models, as shown in the "Source" column of Table~\ref{tab:kitti-stf}. This indicates that our approach effectively maintain accuracy on this source domain while improving performance on the target domain under adverse-weather conditions. 
{We further evaluated the current state-of-the-art voxel-based method GRC~\cite{yang2025towards}, adapted to the range-view setting, for comparison with our approach. Experimental results show that our method still shows significant advantages on SalsaNext compared to GRC, with mIoU increasing by +6.3. The same advantages are also shown on other models. It is worth noting that both RDA~\cite{park2024rethinking} and GRC~\cite{yang2025towards} exhibit varying degrees of performance degradation on the source domain, further highlighting the advantages of our approach.}

\begin{figure*}[th]
    \centering
    \includegraphics[width=\linewidth]{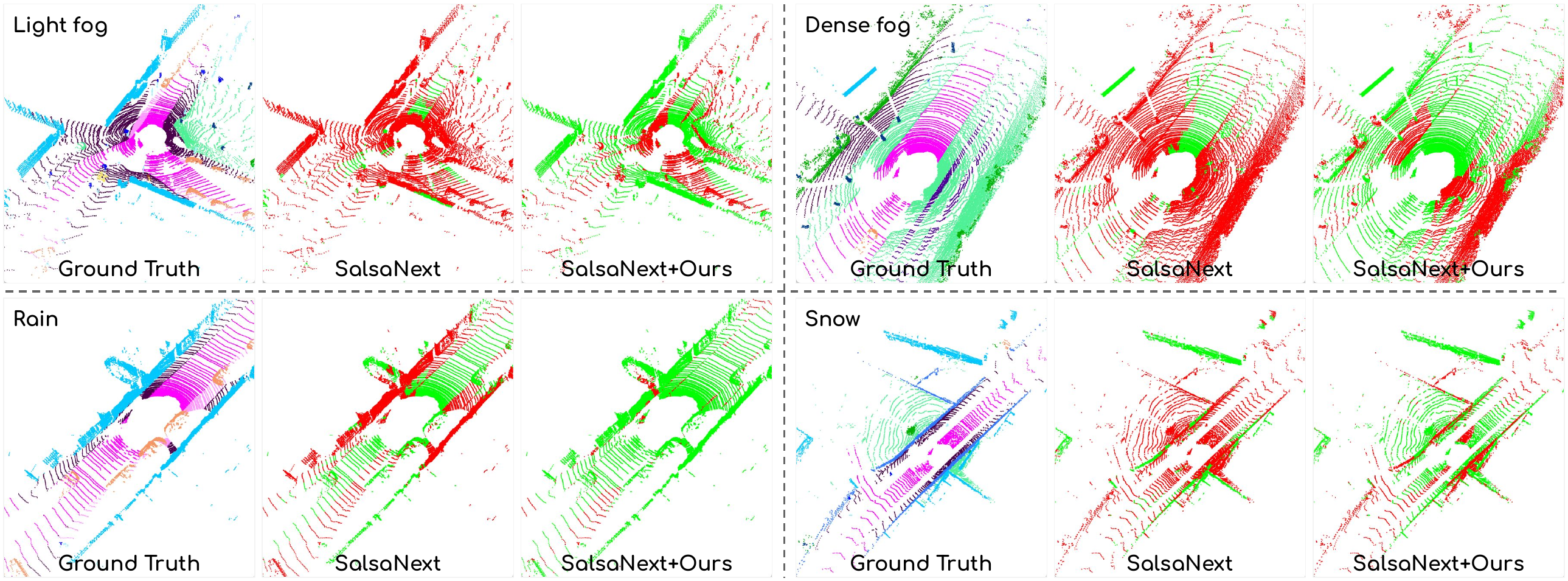}
    \caption{Qualitative results of our method, trained on the SemanticKITTI training set, evaluated under the \textbf{light fog, dense fog, rain and snow} conditions on the SemanticSTF validation set. Green points denote correct predictions, whereas red points indicate incorrect predictions.}
    \label{fig:qualitative_results}
\end{figure*}

Our method is particularly effective at segmenting classes that are challenging for other methods. For instance, in the "car" category, the proposed method achieves a substantial improvement of +46.0 mIoU over the baseline SalsaNext. Similarly, we see notable gains in the "person" category (+41.2 mIoU) and "road" category (+18.9 mIoU), which are typically difficult to segment accurately under adverse conditions. These improvements highlight our method's robustness and its ability to handle both common and safety-critical classes that often suffer from performance degradation in existing models. Furthermore, we notice that applying the augmentation techniques in the existing method PDR~\cite{xiao20233d} leads to a performance decrease on CNN-based models, and very minor improvement for the transformer-based model RangeVIT. In contrast, our approach achieves substantial improvements across most semantic categories. Similar performance improvements are observed across other models, further showcasing the broad applicability and effectiveness of our approach. {Some qualitative results under different weathers are presented in the Figure ~\ref{fig:qualitative_results}.}

\begin{figure}
    \centering
    \includegraphics[width=\linewidth]{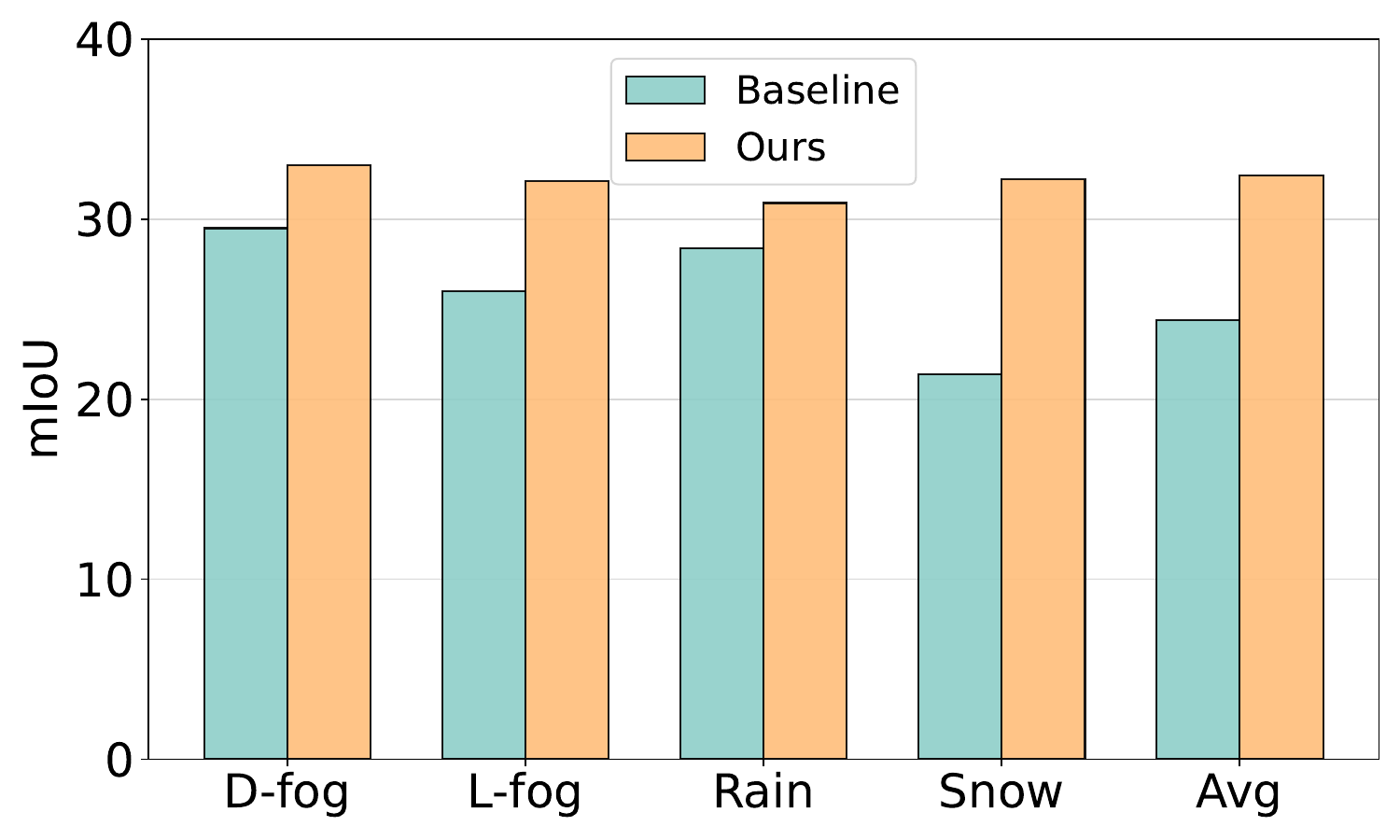}
    \vspace{-0.4cm}
    \caption{Generalized segmentation performance with MinkNet as backbone on SemanticSTF.  }
    \label{fig:minknet}
    \vspace{-0.2cm}
\end{figure}

{In addition, we further verified the effectiveness of our method on voxel-based backbone. As shown in Figure ~\ref{fig:minknet}, we conducted a preliminary study by applying our RDC and GAS modules to a voxel-based MinkNet, and achieved 32.4 mIoU. Compared to the simple MinkNet, this represents an improvement of +7.8 mIoU. Our approach also outperform the baseline in various weather types. While these results do not yet surpass current voxel-based SOTA methods, they demonstrate substantial performance improvements, suggesting that the core ideas behind RDC and GAS are indeed generalizable beyond the range-view setting.}

\vspace{0.3em}
\noindent\textbf{SemanticKITTI to SemanticKITTI-C}. \quad
On SemanticKITTI-C, our method consistently outperforms baseline models. As reported by ``SK$\rightarrow$KC'' in Table~\ref{tab:syn-stf}, CENet improves from 47.7 to 49.3 mIoU, RangeNet++ from 37.1 to 41.7 mIoU, SalsaNext from 40.5 to 49.6 mIoU, and RangeViT from 45.2 to 47.5 mIoU with our approach. These improvements highlight our method's effectiveness in addressing domain shifts and handling corrupted LiDAR data.

\input{table/syn2stf}

\vspace{0.3em}
\noindent\textbf{SynLiDAR to SemanticSTF}. \quad
For the SynLiDAR-to-SemanticSTF experiment, our method also demonstrates a notable improvement across all models compared to the baseline and previous method PDR~\cite{xiao20233d}. As shown by ``Syn$\rightarrow$STF'' of in Table~\ref{tab:syn-stf}, CENet achieves an average mIoU of 4.1, while our method boosts this performance to 9.6, showing a clear gain in handling adverse-weather conditions like dense fog, light fog, rain, and snow. Similarly, RangeNet++ presents an average improvement from 9.3 to 15.3 mIoU, with the significant gains observed under snow (+10.0) and rain (+8.9). SalsaNext also benefits significantly from our approach, with an increase from 9.9 to 14.8 mIoU. RangeViT shows the most dramatic improvement, achieving 17.3 mIoU with our method. Overall, our approach consistently enhances the segmentation performance in all models, surpassing the PointDR method, which shows limited improvement in comparison, particularly in adverse-weather conditions. These results highlight the effectiveness of our method in improving the generalization of range-view-based methods.

\input{table/efficiency}

\vspace{0.3em}
\subsection{Method Analysis}
\noindent\textbf{Efficiency Analysis.}  \quad
Table~\ref{tab:efficiency} presents a detailed comparison of the inference efficiency between our method and the baseline models. Although our approach introduces a very slight overhead, it leads to a significant performance boost under adverse-weather conditions. Across all models, our method achieves a substantial mIoU improvement of at least +17.8, with only a minor increase of approximately 3 ms in latency and a modest addition of about 0.1 M parameters. It is important to note that we do not apply a predefined stem block across all baseline models; rather, we build upon their original stem block designs, which results in slightly varying overheads. For instance, RangeNet++ experiences a more nuanced efficiency impact due to its very lightweight stem block design. While the added latency is minimal (only a 0.7 ms increase), the performance improvement (+19.1 mIoU) highlights the effectiveness of our approach. Our method achieves a significant performance advantage with manageable increases in latency, demonstrating its potential for real-world deployment scenarios where both accuracy and efficiency are critical.

\input{table/ablation}
\vspace{0.3em}

\input{table/resolution}
\vspace{0.3em}

\noindent\textbf{Methodology Ablation.}  \quad
We perform ablation studies with two typical model architectures, including the CNN-based SalsaNext and the Transformer-based RangeViT, to evaluate the effectiveness of the proposed components. The results for each model are presented in Tables~\ref{tab:ablation-salsanext} and~\ref{tab:ablation-rangevit}. 
We take SalsaNext for analysis, the first row shows the baseline model, which exhibits the lowest accuracy under adverse-weather conditions. In the second row, we exclude the reflectance intensity information, using only geometric inputs. This modification results in a significant improvement across all extreme weather conditions, with an average mIoU gain of +12.2 on SemanticSTF. This suggests that the drastic changes in reflectance intensity distribution severely affect the model's robustness. However, this improvement comes at the cost of performance under source domain, with a decrease in mIoU of 1.5 on the source domain.

In the third row, we introduce the Geometric Abnormality Suppression (GAS) module, which further improves performance on the target domain, reaching a mIoU of 24.0. This shows that noise and distortion in the features heavily affect performance, and the GAS module plays a crucial role in mitigating these effects. The fourth row demonstrates that when using both geometric and reflectance features without any additional processing, the model's performance on the SemanticSTF dataset is similar to the baseline, indicating that the inclusion of reflectance alone does not offer generalization benefits.

In the fifth row, we apply the Reflectance Distortion Calibration (RDC) module to the reflectance features. This improves the target domain performance by +9.4 mIoU without negatively impacting performance on the source domain. Finally, the sixth row shows the combination of all the optimization modules, achieving the highest target domain performance of 28.2 mIoU, a substantial improvement over the baseline, with no degradation in source domain performance. 
Experiments on RangeViT also present similar conclusions as shown in Tab.~\ref{tab:ablation-rangevit}, again demonstrating the effectiveness of our designed model.

\noindent\textbf{Impact of Resolution.} \quad
Table~\ref{tab:resolutions} presents the performance of SalsaNext with varying horizontal resolutions for range-view projection. Our proposed method shows consistent improvements across all resolutions. At a resolution of 512, the baseline SalsaNext achieves an average mIoU of 12.5, while our method increases this to 31.0, resulting in a significant improvement of +18.5 mIoU. At 1024 resolution, our method further boosts performance to an average mIoU of 31.4, compared to 9.9 for the baseline, showing a gain of +21.5 mIoU. At the highest resolution, 2048, our method provides a similar improvement.
Increasing the resolution also leads to a rise in latency. At 2048 resolution, the inference time increases to 13.0 ms with our method, compared to 10.1 ms for the baseline. 

We also observe that as resolution increases, the performance under adverse-weather conditions does not consistently improve. In fact, the baseline model shows a noticeable decline in performance with higher resolution. In contrast, performance in the source domain (clean weather) continues to improve as resolution increases. This discrepancy could be attributed to the reduction in point cloud density under adverse-weather conditions; as resolution increases, the proportion of invalid (empty) pixels also rises, reducing the number of valid 2D pixels available for processing.

%% file: table/syn2stf.tex
\begin{table}[t]
    \centering
    \begin{footnotesize}
        \resizebox{0.48\textwidth}{!} {   
    \begin{tabular}{l||cccc|c||c}
        \toprule
        &\multicolumn{ 5 }{c||}{Syn$\rightarrow$STF}& SK$\rightarrow$KC\\
        \hline
        Method & {D-fog} & {L-fog} & {Rain} & {Snow} & {Avg}& {Avg} \\
        \hline
        CENet~\cite{cheng2022cenet} & 6.3 & 3.3 & 4.1 & 2.9 & 4.1 & 47.7 \\
        \quad   w/ PDR~\cite{xiao20233d} & 7.5 & 7.4 & 9.0 & 7.1 & 7.8 & 45.9\\
        \quad  w/ Ours & \textbf{9.7} & \textbf{9.6} & \textbf{12.1} & \textbf{10.0} & \textbf{9.6} & \textbf{49.3}\\
        \hline
        RangeNet$^{\text{++}}$~\cite{milioto2019rangenet++} & 9.9 & 10.8 & 8.7 & 8.2 & 9.3 & 37.1\\
        \quad   w/ PDR~\cite{xiao20233d} & 6.7 & 7.4 & 7.7 & 7.4 & 7.8 & 36.3\\
        \quad  w/ Ours & \textbf{12.8} & \textbf{15.1} & \textbf{17.6} & \textbf{18.2} & \textbf{15.3} & \textbf{41.7}\\
        \hline
        SalsaNext~\cite{cortinhal2020salsanext} & 11.1 & 7.5 & 7.4 & 4.5 & 9.9 &40.5\\
        \quad   w/ PDR~\cite{xiao20233d} & 5.9 & 6.4 & 6.4 & 5.2 & 5.7 &45.5\\
        \quad  w/ Ours & \textbf{12.9} & \textbf{16.1} & \textbf{19.2} & \textbf{15.3} &\textbf{14.8} &\textbf{49.6}\\
        \hline
        RangeViT~\cite{ando2023rangevit} & 10.4 & 7.7 & 8.6 & 9.3 & 8.9 &45.2\\
        \quad   w/ PDR~\cite{xiao20233d} &  4.1 & 3.8 & 2.7 & 2.6 & 3.1 &42.3\\
        \quad  w/ Ours & \textbf{13.9 }& \textbf{17.9} & \textbf{18.7} & \textbf{19.5} & \textbf{17.3} &\textbf{47.5}\\
        \bottomrule
    \end{tabular}
   } 
   \caption{Generalized segmentation performance.  Syn$\rightarrow$STF denotes SynLiDAR$\rightarrow$SemanticSTF and SK$\rightarrow$KC represents SemanticKITTI$\rightarrow$SemanticKITTI-C. }
    \label{tab:syn-stf}  
  \end{footnotesize}
\end{table}

%% file: table/efficiency.tex
\setlength{\tabcolsep}{2.2mm}{
\begin{table}[t]
    \begin{footnotesize}
    \centering       
    \resizebox{0.48\textwidth}{!} {  
    \begin{tabular}{l|cccc}
    \toprule
    Method  & MACs & \#Param & Latency & mIoU \\
      & (G) & (M) & (ms) & (\%) \\
     \hline
    SalsaNext  & 62.8 & 6.7 & 10.1 & 7.9\\
    \quad w/ Ours   & 70.4 & 6.8  & 13.0 & 28.2 \\
     \hline
    RangeNet$^{\text{++}}$  & 377.4 & 50.4 & 17.5 & 7.3\\
    \quad w/ Ours   & 377.4 & 50.4  & 18.2 & 26.4 \\
     \hline
    CENet  & 434.3 & 6.8   & 19.0 & 7.8\\
    \quad w/ Ours    & 463.5 & 7.0  & 23.7 & 25.6 \\
     \hline
    RangeViT  & 404.7 & 27.1  & 49.4 & 10.2\\
    \quad w/ Ours  & 416.1 & 27.2  & 52.7 & 28.9 \\
    \bottomrule
    \end{tabular}
    }    
    \caption{Efficiency Analysis. The latency is measured on a NVIDIA A100 GPU.} 
    \label{tab:efficiency}
    \end{footnotesize}
\end{table}
}

%% file: table/ablation.tex
\setlength{\tabcolsep}{1.4mm}{
\begin{table}[t]
    \begin{footnotesize}
    \centering
    \resizebox{0.48\textwidth}{!} { 
    \begin{tabular}{cccc|ccccc|c}
    \toprule
    &&&&\multicolumn{ 5 }{c|}{SemanticSTF}& Source\\
    Stem$_{G}$  & Stem$_{R}$ & GAS & RDC & D-fog & L-fog & Rain & Snow & Avg & Avg\\
    \hline
                    &             &              &              & 14.3 & 8.1 & 7.0  & 3.9 & 7.9 & 59.7\\
    $\checkmark$    &             &              &              & 20.5 & 21.9 & 20.5 & 19.3 & 20.1 & 58.2\\
    $\checkmark$    &             &$\checkmark$  &              & 23.7 & 27.8 & \textbf{25.1} & \textbf{25.0} & 24.0 &58.3\\
    $\checkmark$    &$\checkmark$ &              &              & 14.5 & 9.6  & 8.9 & 6.5 & 8.8 & 59.7\\
    $\checkmark$    &$\checkmark$ &              &$\checkmark$  & 19.5 & 16.2 & 12.3 & 13.2 & 15.8 & 60.1\\
    $\checkmark$    &$\checkmark$ &$\checkmark$  &$\checkmark$  & \textbf{30.2}  & \textbf{27.9} & 21.5 & 21.9 & \textbf{28.2} & \textbf{60.2}\\
    \bottomrule
    \end{tabular}   
    }
    \caption{Model ablation on SemanticKITTI$\rightarrow$SemanticSTF with SalsaNext as the baseline. Stem$_{G}$ and Stem$_{R}$ represent the two branches, GAS denote the Geometric Abnormality Suppression module, and RDC represents the Reflectance Distortion Calibration module.}
    \label{tab:ablation-salsanext}
    \end{footnotesize}   
\end{table}
}

\setlength{\tabcolsep}{1.4mm}{
\begin{table}[t]
    \begin{footnotesize}
    \centering
    \resizebox{0.48\textwidth}{!} { 
    \begin{tabular}{cccc|ccccc|c}    
    \toprule
    &&&&\multicolumn{ 5 }{c|}{SemanticSTF}& Source\\
    Stem$_{G}$ & Stem$_{R}$ & GAS & RDC & D-fog & L-fog & Rain & Snow & Avg & Avg\\
    \hline
                    &             &              &              & 16.5 & 10.8 & 9.3 & 6.4& 10.2 & 59.6\\
    $\checkmark$    &             &              &              & 16.3 & 17.7 & 16.3 & 15.8 & 16.8 & 59.2\\
    $\checkmark$    &             &$\checkmark$  &              & 20.3 & 21.9 & 21.1 & 17.6 & 19.4 & 58.7\\
    $\checkmark$    &$\checkmark$ &              &              & 17.6 & 13.2 & 8.8 & 4.9 & 10.8 & 60.6\\
    $\checkmark$    &$\checkmark$ &              &$\checkmark$  & 23.2 & 22.2 & 19.1 & 16.9 & 20.2 & \textbf{60.7}\\
    $\checkmark$    &$\checkmark$ &$\checkmark$  &$\checkmark$  & \textbf{29.5} &  \textbf{31.8} & \textbf{30.4} & \textbf{26.2} & \textbf{28.9} & 59.3\\
    \bottomrule
    \end{tabular}
    }
    \caption{Model ablation on SemanticKITTI$\rightarrow$SemanticSTF with RangeViT as the baseline.}
    \label{tab:ablation-rangevit}
    \end{footnotesize}
\end{table}
}

%% file: table/resolution.tex
\setlength{\tabcolsep}{2mm}{
\begin{table*}[t]
    \begin{footnotesize}
    \centering
    \vspace{-0.2cm}
    \resizebox{0.7\textwidth}{!} { 
    \begin{tabular}{c|c|ccccc|c|c}
    \toprule
    &&\multicolumn{ 5 }{c|}{SemanticSTF}& Source &Latency\\
    Method & Res. & D-fog & L-fog & Rain & Snow & Avg & Avg &(ms) \\
    \hline
    SalsaNext & 512 & 5.8 & 7.8 & 11.2 & 12.8 & 12.5 & 57.1 &{6.4}\\
    \quad w/ Ours & 512 & {33.4} & {32.1} & 28.6 & 26.5 & 31.0 & 57.4 &7.8\\
    \hline
    SalsaNext & 1024 & 4.0 & 6.3 & 8.0 & 9.3 & 9.9 & 59.5 &6.8\\
    \quad w/ Ours & 1024 & 30.5 & 31.3 & {28.8} & {30.0} & {31.4} & 59.7  &7.8\\
    \hline
    SalsaNext & 2048 & 14.3 & 8.1 & 7.0  & 3.9 & 7.9 & 59.7 &10.1\\
    \quad w/ Ours & 2048 & 30.2  & 27.9 & 21.5 & 21.9 & 28.2 & {60.2} &13.0\\
    \bottomrule
    \end{tabular}  
    }
    \caption{Performance on SemanticKITTI$\rightarrow$SemanticSTF with different horizontal resolutions for range-view projection.}
    \label{tab:resolutions}
    \end{footnotesize}
    \vspace{-0.cm}
\end{table*}}

%% file: sec/5_conclusion.tex
\section{Conclusion}
In this paper, we propose a novel framework for range-view-based generalized semantic segmentation of LiDAR data, designed to overcome the challenges posed by severe weather conditions. By analyzing the impact of adverse weather on range-view representations, we identify key factors that contribute to performance degradation. Based on this analysis, we introduce a lightweight modular design comprising the Geometric Abnormality Suppression (GAS) module, which dynamically mitigates spatial noise in LiDAR points, and the Reflectance Distortion Calibration (RDC) module, which adaptively corrects distortions in reflectance intensity. Extensive experiments with various baseline models validate the effectiveness and efficiency of our approach, demonstrating significant improvements in generalized segmentation performance under adverse-weather conditions.

%% file: sn-article.bbl

\begin{thebibliography}{49}
\ifx \bisbn   \undefined \def \bisbn  #1{ISBN #1}\fi
\ifx \binits  \undefined \def \binits#1{#1}\fi
\ifx \bauthor  \undefined \def \bauthor#1{#1}\fi
\ifx \batitle  \undefined \def \batitle#1{#1}\fi
\ifx \bjtitle  \undefined \def \bjtitle#1{#1}\fi
\ifx \bvolume  \undefined \def \bvolume#1{\textbf{#1}}\fi
\ifx \byear  \undefined \def \byear#1{#1}\fi
\ifx \bissue  \undefined \def \bissue#1{#1}\fi
\ifx \bfpage  \undefined \def \bfpage#1{#1}\fi
\ifx \blpage  \undefined \def \blpage #1{#1}\fi
\ifx \burl  \undefined \def \burl#1{\textsf{#1}}\fi
\ifx \doiurl  \undefined \def \doiurl#1{\url{https://doi.org/#1}}\fi
\ifx \betal  \undefined \def \betal{\textit{et al.}}\fi
\ifx \binstitute  \undefined \def \binstitute#1{#1}\fi
\ifx \binstitutionaled  \undefined \def \binstitutionaled#1{#1}\fi
\ifx \bctitle  \undefined \def \bctitle#1{#1}\fi
\ifx \beditor  \undefined \def \beditor#1{#1}\fi
\ifx \bpublisher  \undefined \def \bpublisher#1{#1}\fi
\ifx \bbtitle  \undefined \def \bbtitle#1{#1}\fi
\ifx \bedition  \undefined \def \bedition#1{#1}\fi
\ifx \bseriesno  \undefined \def \bseriesno#1{#1}\fi
\ifx \blocation  \undefined \def \blocation#1{#1}\fi
\ifx \bsertitle  \undefined \def \bsertitle#1{#1}\fi
\ifx \bsnm \undefined \def \bsnm#1{#1}\fi
\ifx \bsuffix \undefined \def \bsuffix#1{#1}\fi
\ifx \bparticle \undefined \def \bparticle#1{#1}\fi
\ifx \barticle \undefined \def \barticle#1{#1}\fi
\bibcommenthead
\ifx \bconfdate \undefined \def \bconfdate #1{#1}\fi
\ifx \botherref \undefined \def \botherref #1{#1}\fi
\ifx \url \undefined \def \url#1{\textsf{#1}}\fi
\ifx \bchapter \undefined \def \bchapter#1{#1}\fi
\ifx \bbook \undefined \def \bbook#1{#1}\fi
\ifx \bcomment \undefined \def \bcomment#1{#1}\fi
\ifx \oauthor \undefined \def \oauthor#1{#1}\fi
\ifx \citeauthoryear \undefined \def \citeauthoryear#1{#1}\fi
\ifx \endbibitem  \undefined \def \endbibitem {}\fi
\ifx \bconflocation  \undefined \def \bconflocation#1{#1}\fi
\ifx \arxivurl  \undefined \def \arxivurl#1{\textsf{#1}}\fi
\csname PreBibitemsHook\endcsname

\bibitem[\protect\citeauthoryear{Qi et~al.}{2017}]{qi2017pointnet}
\begin{bchapter}
\bauthor{\bsnm{Qi}, \binits{C.R.}},
\bauthor{\bsnm{Su}, \binits{H.}},
\bauthor{\bsnm{Mo}, \binits{K.}},
\bauthor{\bsnm{Guibas}, \binits{L.J.}}:
\bctitle{Pointnet: Deep learning on point sets for 3d classification and segmentation}.
In: \bbtitle{CVPR}
(\byear{2017})
\end{bchapter}
\endbibitem

\bibitem[\protect\citeauthoryear{Hu et~al.}{2021}]{hu2021learning}
\begin{barticle}
\bauthor{\bsnm{Hu}, \binits{Q.}},
\bauthor{\bsnm{Yang}, \binits{B.}},
\bauthor{\bsnm{Xie}, \binits{L.}},
\bauthor{\bsnm{Rosa}, \binits{S.}},
\bauthor{\bsnm{Guo}, \binits{Y.}},
\bauthor{\bsnm{Wang}, \binits{Z.}},
\bauthor{\bsnm{Trigoni}, \binits{N.}},
\bauthor{\bsnm{Markham}, \binits{A.}}:
\batitle{Learning semantic segmentation of large-scale point clouds with random sampling}.
\bjtitle{IEEE Trans. Pattern Anal. Mach. Intell.}
\bvolume{44}(\bissue{11}),
\bfpage{8338}--\blpage{8354}
(\byear{2021})
\end{barticle}
\endbibitem

\bibitem[\protect\citeauthoryear{Kong et~al.}{2023}]{kong2023rethinking}
\begin{bchapter}
\bauthor{\bsnm{Kong}, \binits{L.}},
\bauthor{\bsnm{Liu}, \binits{Y.}},
\bauthor{\bsnm{Chen}, \binits{R.}},
\bauthor{\bsnm{Ma}, \binits{Y.}},
\bauthor{\bsnm{Zhu}, \binits{X.}},
\bauthor{\bsnm{Li}, \binits{Y.}},
\bauthor{\bsnm{Hou}, \binits{Y.}},
\bauthor{\bsnm{Qiao}, \binits{Y.}},
\bauthor{\bsnm{Liu}, \binits{Z.}}:
\bctitle{Rethinking range view representation for lidar segmentation}.
In: \bbtitle{ICCV}
(\byear{2023})
\end{bchapter}
\endbibitem

\bibitem[\protect\citeauthoryear{Wu et~al.}{2022}]{wu2022point}
\begin{barticle}
\bauthor{\bsnm{Wu}, \binits{X.}},
\bauthor{\bsnm{Lao}, \binits{Y.}},
\bauthor{\bsnm{Jiang}, \binits{L.}},
\bauthor{\bsnm{Liu}, \binits{X.}},
\bauthor{\bsnm{Zhao}, \binits{H.}}:
\batitle{Point transformer v2: Grouped vector attention and partition-based pooling}.
\bjtitle{Adv. Neural Inform. Process. Syst.}
\bvolume{35},
\bfpage{33330}--\blpage{33342}
(\byear{2022})
\end{barticle}
\endbibitem

\bibitem[\protect\citeauthoryear{Wu et~al.}{2024}]{wu2024point}
\begin{bchapter}
\bauthor{\bsnm{Wu}, \binits{X.}},
\bauthor{\bsnm{Jiang}, \binits{L.}},
\bauthor{\bsnm{Wang}, \binits{P.-S.}},
\bauthor{\bsnm{Liu}, \binits{Z.}},
\bauthor{\bsnm{Liu}, \binits{X.}},
\bauthor{\bsnm{Qiao}, \binits{Y.}},
\bauthor{\bsnm{Ouyang}, \binits{W.}},
\bauthor{\bsnm{He}, \binits{T.}},
\bauthor{\bsnm{Zhao}, \binits{H.}}:
\bctitle{Point transformer v3: Simpler faster stronger}.
In: \bbtitle{CVPR}
(\byear{2024})
\end{bchapter}
\endbibitem

\bibitem[\protect\citeauthoryear{Choy et~al.}{2019}]{choy20194d}
\begin{bchapter}
\bauthor{\bsnm{Choy}, \binits{C.}},
\bauthor{\bsnm{Gwak}, \binits{J.}},
\bauthor{\bsnm{Savarese}, \binits{S.}}:
\bctitle{4d spatio-temporal convnets: Minkowski convolutional neural networks}.
In: \bbtitle{CVPR}
(\byear{2019})
\end{bchapter}
\endbibitem

\bibitem[\protect\citeauthoryear{Zhou et~al.}{2021}]{zhou2020cylinder3d}
\begin{bchapter}
\bauthor{\bsnm{Zhou}, \binits{H.}},
\bauthor{\bsnm{Zhu}, \binits{X.}},
\bauthor{\bsnm{Song}, \binits{X.}},
\bauthor{\bsnm{Ma}, \binits{Y.}},
\bauthor{\bsnm{Wang}, \binits{Z.}},
\bauthor{\bsnm{Li}, \binits{H.}},
\bauthor{\bsnm{Lin}, \binits{D.}}:
\bctitle{Cylinder3d: An effective 3d framework for driving-scene lidar semantic segmentation}.
In: \bbtitle{ICCV}
(\byear{2021})
\end{bchapter}
\endbibitem

\bibitem[\protect\citeauthoryear{Yang et~al.}{2023}]{yang2023swin3d}
\begin{botherref}
\oauthor{\bsnm{Yang}, \binits{Y.-Q.}},
\oauthor{\bsnm{Guo}, \binits{Y.-X.}},
\oauthor{\bsnm{Xiong}, \binits{J.-Y.}},
\oauthor{\bsnm{Liu}, \binits{Y.}},
\oauthor{\bsnm{Pan}, \binits{H.}},
\oauthor{\bsnm{Wang}, \binits{P.-S.}},
\oauthor{\bsnm{Tong}, \binits{X.}},
\oauthor{\bsnm{Guo}, \binits{B.}}:
Swin3d: A pretrained transformer backbone for 3d indoor scene understanding.
arXiv preprint arXiv:2304.06906
(2023)
\end{botherref}
\endbibitem

\bibitem[\protect\citeauthoryear{Choe et~al.}{2022}]{choe2022pointmixer}
\begin{bchapter}
\bauthor{\bsnm{Choe}, \binits{J.}},
\bauthor{\bsnm{Park}, \binits{C.}},
\bauthor{\bsnm{Rameau}, \binits{F.}},
\bauthor{\bsnm{Park}, \binits{J.}},
\bauthor{\bsnm{Kweon}, \binits{I.S.}}:
\bctitle{Pointmixer: Mlp-mixer for point cloud understanding}.
In: \bbtitle{ECCV}
(\byear{2022})
\end{bchapter}
\endbibitem

\bibitem[\protect\citeauthoryear{Sun et~al.}{2024}]{sun2024efficient}
\begin{botherref}
\oauthor{\bsnm{Sun}, \binits{X.}},
\oauthor{\bsnm{Liu}, \binits{J.}},
\oauthor{\bsnm{Shen}, \binits{H.T.}},
\oauthor{\bsnm{Zhu}, \binits{X.}},
\oauthor{\bsnm{Hu}, \binits{P.}}:
On efficient variants of segment anything model: A survey.
arXiv preprint arXiv:2410.04960
(2024)
\end{botherref}
\endbibitem

\bibitem[\protect\citeauthoryear{Behley et~al.}{2019}]{behley2019semantickitti}
\begin{bchapter}
\bauthor{\bsnm{Behley}, \binits{J.}},
\bauthor{\bsnm{Garbade}, \binits{M.}},
\bauthor{\bsnm{Milioto}, \binits{A.}},
\bauthor{\bsnm{Quenzel}, \binits{J.}},
\bauthor{\bsnm{Behnke}, \binits{S.}},
\bauthor{\bsnm{Stachniss}, \binits{C.}},
\bauthor{\bsnm{Gall}, \binits{J.}}:
\bctitle{Semantickitti: A dataset for semantic scene understanding of lidar sequences}.
In: \bbtitle{ICCV}
(\byear{2019})
\end{bchapter}
\endbibitem

\bibitem[\protect\citeauthoryear{Caesar et~al.}{2020}]{caesar2020nuscenes}
\begin{bchapter}
\bauthor{\bsnm{Caesar}, \binits{H.}},
\bauthor{\bsnm{Bankiti}, \binits{V.}},
\bauthor{\bsnm{Lang}, \binits{A.H.}},
\bauthor{\bsnm{Vora}, \binits{S.}},
\bauthor{\bsnm{Liong}, \binits{V.E.}},
\bauthor{\bsnm{Xu}, \binits{Q.}},
\bauthor{\bsnm{Krishnan}, \binits{A.}},
\bauthor{\bsnm{Pan}, \binits{Y.}},
\bauthor{\bsnm{Baldan}, \binits{G.}},
\bauthor{\bsnm{Beijbom}, \binits{O.}}:
\bctitle{nuscenes: A multimodal dataset for autonomous driving}.
In: \bbtitle{CVPR}
(\byear{2020})
\end{bchapter}
\endbibitem

\bibitem[\protect\citeauthoryear{Hahner et~al.}{2021}]{hahner2021fog}
\begin{bchapter}
\bauthor{\bsnm{Hahner}, \binits{M.}},
\bauthor{\bsnm{Sakaridis}, \binits{C.}},
\bauthor{\bsnm{Dai}, \binits{D.}},
\bauthor{\bsnm{Van~Gool}, \binits{L.}}:
\bctitle{Fog simulation on real lidar point clouds for 3d object detection in adverse weather}.
In: \bbtitle{ICCV}
(\byear{2021})
\end{bchapter}
\endbibitem

\bibitem[\protect\citeauthoryear{Bijelic et~al.}{2020}]{bijelic2020seeing}
\begin{bchapter}
\bauthor{\bsnm{Bijelic}, \binits{M.}},
\bauthor{\bsnm{Gruber}, \binits{T.}},
\bauthor{\bsnm{Mannan}, \binits{F.}},
\bauthor{\bsnm{Kraus}, \binits{F.}},
\bauthor{\bsnm{Ritter}, \binits{W.}},
\bauthor{\bsnm{Dietmayer}, \binits{K.}},
\bauthor{\bsnm{Heide}, \binits{F.}}:
\bctitle{Seeing through fog without seeing fog: Deep multimodal sensor fusion in unseen adverse weather}.
In: \bbtitle{CVPR}
(\byear{2020})
\end{bchapter}
\endbibitem

\bibitem[\protect\citeauthoryear{Xiao et~al.}{2023}]{xiao20233d}
\begin{bchapter}
\bauthor{\bsnm{Xiao}, \binits{A.}},
\bauthor{\bsnm{Huang}, \binits{J.}},
\bauthor{\bsnm{Xuan}, \binits{W.}},
\bauthor{\bsnm{Ren}, \binits{R.}},
\bauthor{\bsnm{Liu}, \binits{K.}},
\bauthor{\bsnm{Guan}, \binits{D.}},
\bauthor{\bsnm{El~Saddik}, \binits{A.}},
\bauthor{\bsnm{Lu}, \binits{S.}},
\bauthor{\bsnm{Xing}, \binits{E.P.}}:
\bctitle{3d semantic segmentation in the wild: Learning generalized models for adverse-condition point clouds}.
In: \bbtitle{CVPR}
(\byear{2023})
\end{bchapter}
\endbibitem

\bibitem[\protect\citeauthoryear{Hahner et~al.}{2022}]{hahner2022lidar}
\begin{bchapter}
\bauthor{\bsnm{Hahner}, \binits{M.}},
\bauthor{\bsnm{Sakaridis}, \binits{C.}},
\bauthor{\bsnm{Bijelic}, \binits{M.}},
\bauthor{\bsnm{Heide}, \binits{F.}},
\bauthor{\bsnm{Yu}, \binits{F.}},
\bauthor{\bsnm{Dai}, \binits{D.}},
\bauthor{\bsnm{Van~Gool}, \binits{L.}}:
\bctitle{Lidar snowfall simulation for robust 3d object detection}.
In: \bbtitle{CVPR}
(\byear{2022})
\end{bchapter}
\endbibitem

\bibitem[\protect\citeauthoryear{Zhao et~al.}{2024}]{zhao2024unimix}
\begin{bchapter}
\bauthor{\bsnm{Zhao}, \binits{H.}},
\bauthor{\bsnm{Zhang}, \binits{J.}},
\bauthor{\bsnm{Chen}, \binits{Z.}},
\bauthor{\bsnm{Zhao}, \binits{S.}},
\bauthor{\bsnm{Tao}, \binits{D.}}:
\bctitle{Unimix: Towards domain adaptive and generalizable lidar semantic segmentation in adverse weather}.
In: \bbtitle{CVPR}
(\byear{2024})
\end{bchapter}
\endbibitem

\bibitem[\protect\citeauthoryear{Yang et~al.}{2023}]{yang2023realistic}
\begin{botherref}
\oauthor{\bsnm{Yang}, \binits{D.}},
\oauthor{\bsnm{Liu}, \binits{Z.}},
\oauthor{\bsnm{Jiang}, \binits{W.}},
\oauthor{\bsnm{Yan}, \binits{G.}},
\oauthor{\bsnm{Gao}, \binits{X.}},
\oauthor{\bsnm{Shi}, \binits{B.}},
\oauthor{\bsnm{Liu}, \binits{S.}},
\oauthor{\bsnm{Cai}, \binits{X.}}:
Realistic rainy weather simulation for lidars in carla simulator.
arXiv preprint arXiv:2312.12772
(2023)
\end{botherref}
\endbibitem

\bibitem[\protect\citeauthoryear{Park et~al.}{2024}]{park2024rethinking}
\begin{botherref}
\oauthor{\bsnm{Park}, \binits{J.}},
\oauthor{\bsnm{Kim}, \binits{K.}},
\oauthor{\bsnm{Shim}, \binits{H.}}:
Rethinking data augmentation for robust lidar semantic segmentation in adverse weather.
arXiv preprint arXiv:2407.02286
(2024)
\end{botherref}
\endbibitem

\bibitem[\protect\citeauthoryear{He et~al.}{2024}]{he2024domain}
\begin{bchapter}
\bauthor{\bsnm{He}, \binits{P.}},
\bauthor{\bsnm{Jiao}, \binits{L.}},
\bauthor{\bsnm{Li}, \binits{L.}},
\bauthor{\bsnm{Liu}, \binits{X.}},
\bauthor{\bsnm{Liu}, \binits{F.}},
\bauthor{\bsnm{Ma}, \binits{W.}},
\bauthor{\bsnm{Yang}, \binits{S.}},
\bauthor{\bsnm{Shang}, \binits{R.}}:
\bctitle{Domain generalization-aware uncertainty introspective learning for 3d point clouds segmentation}.
In: \bbtitle{ACMMM}
(\byear{2024})
\end{bchapter}
\endbibitem

\bibitem[\protect\citeauthoryear{Park et~al.}{2025}]{park2025no}
\begin{bchapter}
\bauthor{\bsnm{Park}, \binits{J.}},
\bauthor{\bsnm{Lee}, \binits{H.}},
\bauthor{\bsnm{Kang}, \binits{I.}},
\bauthor{\bsnm{Shim}, \binits{H.}}:
\bctitle{No thing, nothing: Highlighting safety-critical classes for robust lidar semantic segmentation in adverse weather}.
In: \bbtitle{Proceedings of the Computer Vision and Pattern Recognition Conference},
pp. \bfpage{6690}--\blpage{6699}
(\byear{2025})
\end{bchapter}
\endbibitem

\bibitem[\protect\citeauthoryear{Qi et~al.}{2017}]{qi2017pointnet++}
\begin{botherref}
\oauthor{\bsnm{Qi}, \binits{C.R.}},
\oauthor{\bsnm{Yi}, \binits{L.}},
\oauthor{\bsnm{Su}, \binits{H.}},
\oauthor{\bsnm{Guibas}, \binits{L.J.}}:
Pointnet++: Deep hierarchical feature learning on point sets in a metric space.
Adv. Neural Inform. Process. Syst.
\textbf{30}
(2017)
\end{botherref}
\endbibitem

\bibitem[\protect\citeauthoryear{Thomas et~al.}{2019}]{thomas2019kpconv}
\begin{bchapter}
\bauthor{\bsnm{Thomas}, \binits{H.}},
\bauthor{\bsnm{Qi}, \binits{C.R.}},
\bauthor{\bsnm{Deschaud}, \binits{J.-E.}},
\bauthor{\bsnm{Marcotegui}, \binits{B.}},
\bauthor{\bsnm{Goulette}, \binits{F.}},
\bauthor{\bsnm{Guibas}, \binits{L.J.}}:
\bctitle{Kpconv: Flexible and deformable convolution for point clouds}.
In: \bbtitle{ICCV}
(\byear{2019})
\end{bchapter}
\endbibitem

\bibitem[\protect\citeauthoryear{Wu et~al.}{2019}]{wu2019pointconv}
\begin{bchapter}
\bauthor{\bsnm{Wu}, \binits{W.}},
\bauthor{\bsnm{Qi}, \binits{Z.}},
\bauthor{\bsnm{Fuxin}, \binits{L.}}:
\bctitle{Pointconv: Deep convolutional networks on 3d point clouds}.
In: \bbtitle{CVPR}
(\byear{2019})
\end{bchapter}
\endbibitem

\bibitem[\protect\citeauthoryear{Liu et~al.}{2020}]{liu2020closer}
\begin{bchapter}
\bauthor{\bsnm{Liu}, \binits{Z.}},
\bauthor{\bsnm{Hu}, \binits{H.}},
\bauthor{\bsnm{Cao}, \binits{Y.}},
\bauthor{\bsnm{Zhang}, \binits{Z.}},
\bauthor{\bsnm{Tong}, \binits{X.}}:
\bctitle{A closer look at local aggregation operators in point cloud analysis}.
In: \bbtitle{Computer Vision--ECCV 2020: 16th European Conference, Glasgow, UK, August 23--28, 2020, Proceedings, Part XXIII 16},
pp. \bfpage{326}--\blpage{342}
(\byear{2020}).
\bcomment{Springer}
\end{bchapter}
\endbibitem

\bibitem[\protect\citeauthoryear{Wang et~al.}{2019}]{wang2019dynamic}
\begin{barticle}
\bauthor{\bsnm{Wang}, \binits{Y.}},
\bauthor{\bsnm{Sun}, \binits{Y.}},
\bauthor{\bsnm{Liu}, \binits{Z.}},
\bauthor{\bsnm{Sarma}, \binits{S.E.}},
\bauthor{\bsnm{Bronstein}, \binits{M.M.}},
\bauthor{\bsnm{Solomon}, \binits{J.M.}}:
\batitle{Dynamic graph cnn for learning on point clouds}.
\bjtitle{ACM Trans. Graph.}
\bvolume{38}(\bissue{5}),
\bfpage{1}--\blpage{12}
(\byear{2019})
\end{barticle}
\endbibitem

\bibitem[\protect\citeauthoryear{Landrieu and Simonovsky}{2018}]{landrieu2018large}
\begin{bchapter}
\bauthor{\bsnm{Landrieu}, \binits{L.}},
\bauthor{\bsnm{Simonovsky}, \binits{M.}}:
\bctitle{Large-scale point cloud semantic segmentation with superpoint graphs}.
In: \bbtitle{CVPR}
(\byear{2018})
\end{bchapter}
\endbibitem

\bibitem[\protect\citeauthoryear{Du et~al.}{2022}]{du2022novel}
\begin{barticle}
\bauthor{\bsnm{Du}, \binits{Z.}},
\bauthor{\bsnm{Ye}, \binits{H.}},
\bauthor{\bsnm{Cao}, \binits{F.}}:
\batitle{A novel local--global graph convolutional method for point cloud semantic segmentation}.
\bjtitle{IEEE Transactions on Neural Networks and Learning Systems}
\bvolume{35}(\bissue{4}),
\bfpage{4798}--\blpage{4812}
(\byear{2022})
\end{barticle}
\endbibitem

\bibitem[\protect\citeauthoryear{Zhao et~al.}{2021}]{zhao2021point}
\begin{bchapter}
\bauthor{\bsnm{Zhao}, \binits{H.}},
\bauthor{\bsnm{Jiang}, \binits{L.}},
\bauthor{\bsnm{Jia}, \binits{J.}},
\bauthor{\bsnm{Torr}, \binits{P.H.}},
\bauthor{\bsnm{Koltun}, \binits{V.}}:
\bctitle{Point transformer}.
In: \bbtitle{ICCV}
(\byear{2021})
\end{bchapter}
\endbibitem

\bibitem[\protect\citeauthoryear{Lai et~al.}{2022}]{lai2022stratified}
\begin{bchapter}
\bauthor{\bsnm{Lai}, \binits{X.}},
\bauthor{\bsnm{Liu}, \binits{J.}},
\bauthor{\bsnm{Jiang}, \binits{L.}},
\bauthor{\bsnm{Wang}, \binits{L.}},
\bauthor{\bsnm{Zhao}, \binits{H.}},
\bauthor{\bsnm{Liu}, \binits{S.}},
\bauthor{\bsnm{Qi}, \binits{X.}},
\bauthor{\bsnm{Jia}, \binits{J.}}:
\bctitle{Stratified transformer for 3d point cloud segmentation}.
In: \bbtitle{CVPR}
(\byear{2022})
\end{bchapter}
\endbibitem

\bibitem[\protect\citeauthoryear{Robert et~al.}{2023}]{robert2023efficient}
\begin{botherref}
\oauthor{\bsnm{Robert}, \binits{D.}},
\oauthor{\bsnm{Raguet}, \binits{H.}},
\oauthor{\bsnm{Landrieu}, \binits{L.}}:
Efficient 3d semantic segmentation with superpoint transformer. arxiv.
arXiv preprint arXiv:2306.08045
(2023)
\end{botherref}
\endbibitem

\bibitem[\protect\citeauthoryear{Wu et~al.}{2015}]{wu20153d}
\begin{bchapter}
\bauthor{\bsnm{Wu}, \binits{Z.}},
\bauthor{\bsnm{Song}, \binits{S.}},
\bauthor{\bsnm{Khosla}, \binits{A.}},
\bauthor{\bsnm{Yu}, \binits{F.}},
\bauthor{\bsnm{Zhang}, \binits{L.}},
\bauthor{\bsnm{Tang}, \binits{X.}},
\bauthor{\bsnm{Xiao}, \binits{J.}}:
\bctitle{3d shapenets: A deep representation for volumetric shapes}.
In: \bbtitle{CVPR}
(\byear{2015})
\end{bchapter}
\endbibitem

\bibitem[\protect\citeauthoryear{Lai et~al.}{2023}]{lai2023spherical}
\begin{bchapter}
\bauthor{\bsnm{Lai}, \binits{X.}},
\bauthor{\bsnm{Chen}, \binits{Y.}},
\bauthor{\bsnm{Lu}, \binits{F.}},
\bauthor{\bsnm{Liu}, \binits{J.}},
\bauthor{\bsnm{Jia}, \binits{J.}}:
\bctitle{Spherical transformer for lidar-based 3d recognition}.
In: \bbtitle{CVPR}
(\byear{2023})
\end{bchapter}
\endbibitem

\bibitem[\protect\citeauthoryear{Han et~al.}{2020}]{han2020occuseg}
\begin{bchapter}
\bauthor{\bsnm{Han}, \binits{L.}},
\bauthor{\bsnm{Zheng}, \binits{T.}},
\bauthor{\bsnm{Xu}, \binits{L.}},
\bauthor{\bsnm{Fang}, \binits{L.}}:
\bctitle{Occuseg: Occupancy-aware 3d instance segmentation}.
In: \bbtitle{Proceedings of the IEEE/CVF Conference on Computer Vision and Pattern Recognition},
pp. \bfpage{2940}--\blpage{2949}
(\byear{2020})
\end{bchapter}
\endbibitem

\bibitem[\protect\citeauthoryear{Graham et~al.}{2018}]{graham20183d}
\begin{bchapter}
\bauthor{\bsnm{Graham}, \binits{B.}},
\bauthor{\bsnm{Engelcke}, \binits{M.}},
\bauthor{\bsnm{Van Der~Maaten}, \binits{L.}}:
\bctitle{3d semantic segmentation with submanifold sparse convolutional networks}.
In: \bbtitle{CVPR}
(\byear{2018})
\end{bchapter}
\endbibitem

\bibitem[\protect\citeauthoryear{Cortinhal et~al.}{2020}]{cortinhal2020salsanext}
\begin{bchapter}
\bauthor{\bsnm{Cortinhal}, \binits{T.}},
\bauthor{\bsnm{Tzelepis}, \binits{G.}},
\bauthor{\bsnm{Erdal~Aksoy}, \binits{E.}}:
\bctitle{Salsanext: Fast, uncertainty-aware semantic segmentation of lidar point clouds}.
In: \bbtitle{ISVC}
(\byear{2020})
\end{bchapter}
\endbibitem

\bibitem[\protect\citeauthoryear{Milioto et~al.}{2019}]{milioto2019rangenet++}
\begin{bchapter}
\bauthor{\bsnm{Milioto}, \binits{A.}},
\bauthor{\bsnm{Vizzo}, \binits{I.}},
\bauthor{\bsnm{Behley}, \binits{J.}},
\bauthor{\bsnm{Stachniss}, \binits{C.}}:
\bctitle{Rangenet++: Fast and accurate lidar semantic segmentation}.
In: \bbtitle{IROS}
(\byear{2019})
\end{bchapter}
\endbibitem

\bibitem[\protect\citeauthoryear{Li et~al.}{2025}]{li2025rapid}
\begin{bchapter}
\bauthor{\bsnm{Li}, \binits{L.}},
\bauthor{\bsnm{Shum}, \binits{H.P.}},
\bauthor{\bsnm{Breckon}, \binits{T.P.}}:
\bctitle{Rapid-seg: Range-aware pointwise distance distribution networks for 3d lidar segmentation}.
In: \bbtitle{ECCV}
(\byear{2025})
\end{bchapter}
\endbibitem

\bibitem[\protect\citeauthoryear{Ando et~al.}{2023}]{ando2023rangevit}
\begin{bchapter}
\bauthor{\bsnm{Ando}, \binits{A.}},
\bauthor{\bsnm{Gidaris}, \binits{S.}},
\bauthor{\bsnm{Bursuc}, \binits{A.}},
\bauthor{\bsnm{Puy}, \binits{G.}},
\bauthor{\bsnm{Boulch}, \binits{A.}},
\bauthor{\bsnm{Marlet}, \binits{R.}}:
\bctitle{Rangevit: Towards vision transformers for 3d semantic segmentation in autonomous driving}.
In: \bbtitle{CVPR}
(\byear{2023})
\end{bchapter}
\endbibitem

\bibitem[\protect\citeauthoryear{Cheng et~al.}{2022}]{cheng2022cenet}
\begin{bchapter}
\bauthor{\bsnm{Cheng}, \binits{H.-X.}},
\bauthor{\bsnm{Han}, \binits{X.-F.}},
\bauthor{\bsnm{Xiao}, \binits{G.-Q.}}:
\bctitle{Cenet: Toward concise and efficient lidar semantic segmentation for autonomous driving}.
In: \bbtitle{2022 IEEE International Conference on Multimedia and Expo (ICME)},
pp. \bfpage{01}--\blpage{06}
(\byear{2022}).
\bcomment{IEEE}
\end{bchapter}
\endbibitem

\bibitem[\protect\citeauthoryear{Yan et~al.}{2024}]{yan2024benchmarking}
\begin{botherref}
\oauthor{\bsnm{Yan}, \binits{X.}},
\oauthor{\bsnm{Zheng}, \binits{C.}},
\oauthor{\bsnm{Xue}, \binits{Y.}},
\oauthor{\bsnm{Li}, \binits{Z.}},
\oauthor{\bsnm{Cui}, \binits{S.}},
\oauthor{\bsnm{Dai}, \binits{D.}}:
Benchmarking the robustness of lidar semantic segmentation models.
Int. J. Comput. Vis.,
1--24
(2024)
\end{botherref}
\endbibitem

\bibitem[\protect\citeauthoryear{Kong et~al.}{2023}]{kong2023robo3d}
\begin{bchapter}
\bauthor{\bsnm{Kong}, \binits{L.}},
\bauthor{\bsnm{Liu}, \binits{Y.}},
\bauthor{\bsnm{Li}, \binits{X.}},
\bauthor{\bsnm{Chen}, \binits{R.}},
\bauthor{\bsnm{Zhang}, \binits{W.}},
\bauthor{\bsnm{Ren}, \binits{J.}},
\bauthor{\bsnm{Pan}, \binits{L.}},
\bauthor{\bsnm{Chen}, \binits{K.}},
\bauthor{\bsnm{Liu}, \binits{Z.}}:
\bctitle{Robo3d: Towards robust and reliable 3d perception against corruptions}.
In: \bbtitle{ICCV}
(\byear{2023})
\end{bchapter}
\endbibitem

\bibitem[\protect\citeauthoryear{Drake and Gordon}{1985}]{drake1985mie}
\begin{barticle}
\bauthor{\bsnm{Drake}, \binits{R.}},
\bauthor{\bsnm{Gordon}, \binits{J.}}:
\batitle{Mie scattering}.
\bjtitle{American Journal of Physics}
\bvolume{53}(\bissue{10}),
\bfpage{955}--\blpage{962}
(\byear{1985})
\end{barticle}
\endbibitem

\bibitem[\protect\citeauthoryear{Hojjati et~al.}{2024}]{hojjati2024self}
\begin{barticle}
\bauthor{\bsnm{Hojjati}, \binits{H.}},
\bauthor{\bsnm{Ho}, \binits{T.K.K.}},
\bauthor{\bsnm{Armanfard}, \binits{N.}}:
\batitle{Self-supervised anomaly detection in computer vision and beyond: A survey and outlook}.
\bjtitle{Neural Networks}
\bvolume{172},
\bfpage{106106}
(\byear{2024})
\end{barticle}
\endbibitem

\bibitem[\protect\citeauthoryear{Ruff et~al.}{2021}]{ruff2021unifying}
\begin{barticle}
\bauthor{\bsnm{Ruff}, \binits{L.}},
\bauthor{\bsnm{Kauffmann}, \binits{J.R.}},
\bauthor{\bsnm{Vandermeulen}, \binits{R.A.}},
\bauthor{\bsnm{Montavon}, \binits{G.}},
\bauthor{\bsnm{Samek}, \binits{W.}},
\bauthor{\bsnm{Kloft}, \binits{M.}},
\bauthor{\bsnm{Dietterich}, \binits{T.G.}},
\bauthor{\bsnm{M{\"u}ller}, \binits{K.-R.}}:
\batitle{A unifying review of deep and shallow anomaly detection}.
\bjtitle{Proceedings of the IEEE}
\bvolume{109}(\bissue{5}),
\bfpage{756}--\blpage{795}
(\byear{2021})
\end{barticle}
\endbibitem

\bibitem[\protect\citeauthoryear{Huang and Belongie}{2017}]{huang2017arbitrary}
\begin{bchapter}
\bauthor{\bsnm{Huang}, \binits{X.}},
\bauthor{\bsnm{Belongie}, \binits{S.}}:
\bctitle{Arbitrary style transfer in real-time with adaptive instance normalization}.
In: \bbtitle{ICCV}
(\byear{2017})
\end{bchapter}
\endbibitem

\bibitem[\protect\citeauthoryear{Yang et~al.}{2025}]{yang2025towards}
\begin{bchapter}
\bauthor{\bsnm{Yang}, \binits{L.}},
\bauthor{\bsnm{Hu}, \binits{P.}},
\bauthor{\bsnm{Yuan}, \binits{S.}},
\bauthor{\bsnm{Zhang}, \binits{L.}},
\bauthor{\bsnm{Liu}, \binits{J.}},
\bauthor{\bsnm{Shen}, \binits{H.}},
\bauthor{\bsnm{Zhu}, \binits{X.}}:
\bctitle{Towards explicit geometry-reflectance collaboration for generalized lidar segmentation in adverse weather}.
In: \bbtitle{CVPR}
(\byear{2025})
\end{bchapter}
\endbibitem

\bibitem[\protect\citeauthoryear{Geiger et~al.}{2012}]{geiger2012we}
\begin{bchapter}
\bauthor{\bsnm{Geiger}, \binits{A.}},
\bauthor{\bsnm{Lenz}, \binits{P.}},
\bauthor{\bsnm{Urtasun}, \binits{R.}}:
\bctitle{Are we ready for autonomous driving? the kitti vision benchmark suite}.
In: \bbtitle{CVPR}
(\byear{2012})
\end{bchapter}
\endbibitem

\bibitem[\protect\citeauthoryear{Xiao et~al.}{2022}]{xiao2022transfer}
\begin{bchapter}
\bauthor{\bsnm{Xiao}, \binits{A.}},
\bauthor{\bsnm{Huang}, \binits{J.}},
\bauthor{\bsnm{Guan}, \binits{D.}},
\bauthor{\bsnm{Zhan}, \binits{F.}},
\bauthor{\bsnm{Lu}, \binits{S.}}:
\bctitle{Transfer learning from synthetic to real lidar point cloud for semantic segmentation}.
In: \bbtitle{AAAI}
(\byear{2022})
\end{bchapter}
\endbibitem

\end{thebibliography}
